\documentclass{article}

\usepackage{microtype}
\usepackage{graphicx}
\usepackage{booktabs} 

\usepackage{hyperref}

\usepackage[accepted]{icml2024}

\usepackage{amsmath}
\usepackage{amssymb}
\usepackage{mathtools}
\usepackage{amsthm}

\usepackage[capitalize,noabbrev]{cleveref}

\usepackage[justification=centering]{caption}
\usepackage{booktabs}
\usepackage{multirow}
\usepackage[normalem]{ulem}
\useunder{\uline}{\ul}{}
\usepackage{threeparttable}
\usepackage{algorithm}
\usepackage{adjustbox} 
\usepackage{subfig} 

\theoremstyle{plain}

\theoremstyle{definition}

\theoremstyle{remark}

\usepackage[textsize=tiny]{todonotes}

\icmltitlerunning{}
\begin{document}
\twocolumn[
\icmltitle{EasyFS: an Efficient Model-free Feature Selection Framework \\via Elastic Transformation of Features}




\begin{icmlauthorlist}
\icmlauthor{Jianming Lv}{yyy}
\icmlauthor{Sijun Xia}{yyy}
\icmlauthor{Depin Liang}{yyy}
\icmlauthor{Wei Chen}{sch}
\end{icmlauthorlist}

\icmlaffiliation{yyy}{South China University of Technology,Guangzhou Guangdong, China}
\icmlaffiliation{sch}{Chinese Academy of Sciences, Beijing, China}

\icmlcorrespondingauthor{Jianming Lv}{jmlv@scut.edu.cn}

\icmlkeywords{Machine Learning, ICML}

\vskip 0.3in
]



\printAffiliationsAndNotice{}  

\begin{abstract}
Traditional model-free feature selection methods treat each feature independently while disregarding the interrelationships among features,  which leads to relatively poor performance compared with the model-aware methods. To address this challenge, we propose an efficient model-free feature selection framework via  elastic expansion and compression of the features,  namely EasyFS,  to achieve better performance than state-of-the-art model-aware methods while sharing the characters of efficiency and flexibility with the existing model-free methods. In particular, EasyFS expands the feature space by using the random non-linear projection  network to achieve the non-linear combinations of the original features, so as to model the interrelationships among the features and discover most correlated features. Meanwhile, a novel redundancy measurement based on the change of coding rate is proposed for efficient filtering of redundant features. Comprehensive experiments on 21 different datasets show that EasyFS outperforms state-of-the art methods up to 10.9\% in the regression tasks and 5.7\% in the classification tasks while saving more than 94\% of the time.

\end{abstract}
\section{Introduction}
With the rapid development of the Internet and modern industrialization, various industries are acquiring and accumulating data at an unprecedented pace \cite{yin2014review}.  Despite the fast-growing data collection process, researchers face severe challenges when processing massive data that is inundated with numerous high-dimensional features, which significantly increases computational costs and leads to overfitting problems \cite{jia2022feature}. As the key pre-processing procedure, feature  selection aims to  reduce the number of features, so as to  reduce the timing cost and also enhance the interpretability of algorithms. In particular, by eliminating redundant or irrelevant input features, it allows models to more efficiently utilize data, thereby increasing users' trust in the model's predictive results and addressing the question of `why the algorithm is effective\cite{moradipari2022feature}.  

According to the dependency of the downstream classifier/regression model, existing feature selection techniques can be typically categorized into two types: model-aware and model-free methods. The model-aware methods optimize the features based on the downstream learning models. In particular, some genetic algorithm based research \cite{xue2018novel} \cite{chen2013efficient} treat the learning model as a black box, and select the features which  can lead to best performance based on the evaluation metrics provided by the downstream model. Another kind of model-aware methods (e.g., DFS\cite{li2016deep},STG\cite{stg},LassoNet\cite{lassonet}, etc.) incorporate the feature selection process into the optimization process of learning models, and rely on the gradient descent optimization of the whole model with multiple rounds of iterative training. Most of the model-aware methods tend to be time-consuming, because the running and optimization of the downstream models are necessary, which usually adopt deep neural network architectures for high performance. Meanwhile, these feature selection procedures bind with specific downstream model, which is not flexible for further model selection. 

Different from the model-aware methods, the model-free methods \cite{relif}\cite{fisher_score}\cite{CIFE}\cite{DISR}\cite{ifs}  are usually characterized by their lower computing cost and higher flexibility, which select the features with no need for any prior knowledge of downstream learning models. These methods typically  rank the features according to the intrinsic properties of individual features (e.g., target correlation, variance, locality, information gain, etc.). While the features are usually correlated with each other in the practical complex use cases with high-dimensional feature space, ignoring the effects of feature interactions and fusion between features usually leads to weaker performance of existing model-free methods compared with the model-aware ones.

In this paper, we try to propose an efficient model-free feature selection framework based on the random projecting network, namely EasyFS, to achieve the following goals simultaneously: effectiveness, efficiency, and flexibility. In particular, the elastic transformation of features in EasyFS contains two opposite transformation procedures: feature extension and feature compression. In the feature extension stage, EasyFS introduces the interaction between the features by applying the lightweight random projecting network to extend the feature space with non-linear combinations of original features. In the feature compression stage, both relevance and redundancy of  the features are considered to select the most correlated and less redundant features. Specifically, the redundancy of the features are efficiently measured by the change of coding rate, and deployed to filter out the less useful features. Compared with the traditional entropy-based redundancy measurement \cite{mrmr}, which does not naturally support continuous variables and is relatively slow in computation, our proposed coding rate based measurement in a compact matrix form is easy to be accelerated with parallel matrix manipulation. Furthermore, with no need of invoking the downstream learning models, EasyFS is totally model-free and much more efficient than the model-aware methods, while even achieving better performance in the downstream tasks.
The contributions are summarized as follows:
\begin{itemize}
\item{We propose a novel model-free feature selection framework based on the elastic transformation of features, namely EasyFS, which utilizes the lightweight random projection network to expand the original features, so as to extract the deep relationships between features.} 
\item{We propose a efficient method to calculate the redundancy of features, which characterizes the global redundancy of features by computing the variation in coding rate of feature matrix after the removal of a specific feature.  In comparison to traditional mutual information based redundancy measurement, this approach naturally supports continuous variables and is faster in computation.}
\item{We have investigated the feature selection problems for both regression and classification tasks to verify the generalization capability of EasyFS. Comprehensive experiments on 21 tasks are conducted to show that EasyFS outperforms state-of-the art methods up to 10.9\% in the regression tasks and 5.7\% in the classification tasks while saving more than 94\% of the time.} 
\end{itemize}




\begin{figure*}[ht]
\vskip 0.2in
\begin{center}
\centerline{\includegraphics[width=1.0\textwidth]{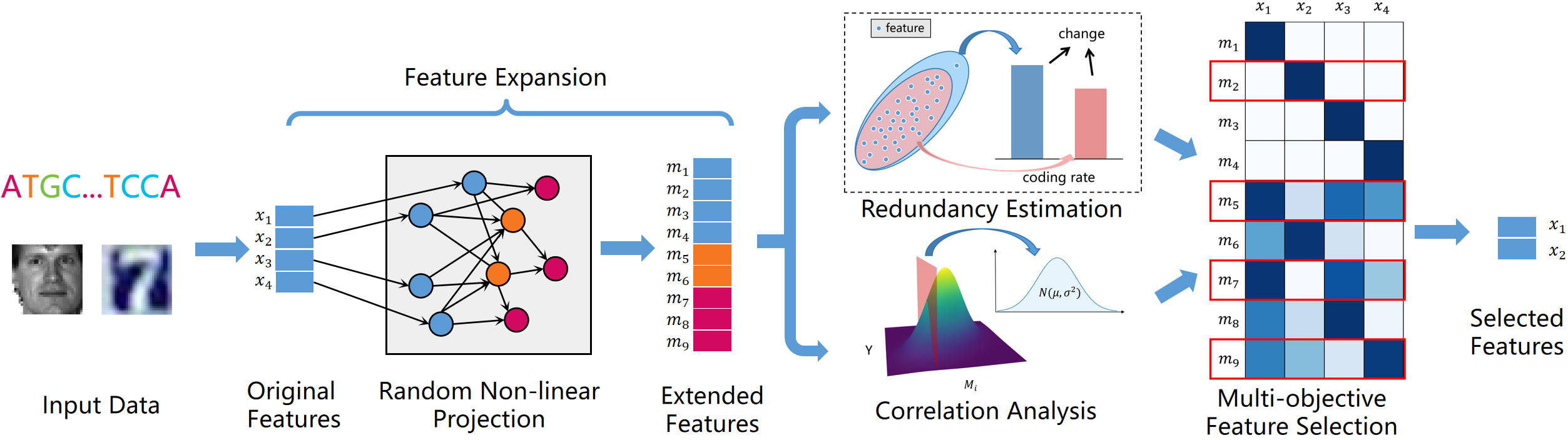}}
\caption{The pipe-line of EasyFS}
\label{fig:overview}
\end{center}
\vskip -0.2in
\end{figure*}

 \section{Related Works}
Supervised feature selection methods can be divided into model-aware and model-free methods roughly, based on whether there is a need to explicitly define the downstream learner.


In particular, the model-aware methods include the wrapper and embedded approaches according to {\color{black} \cite{zebari2020comprehensive}}. The wrapper methods treat the learner as a black box. During each selection iteration, a specified number of features are selected and replaced based on the evaluation metrics provided by the downstream learner, so as to test and identify a better feature subset. E.g. HGEFS\cite{xue2018novel} performed feature selection by combining a genetic algorithm with the Extreme Learning Machine (ELM). The Ant Colony Optimization (ACO) has also been used for feature selection in image processing \cite{chen2013efficient}. All of above methods rely on the running of the downstream predictor, which makes the feature selection procedure time-consuming. Additionally, re-learning of the optimized features is usually required when the downstream learning models are changed, and these methods are prone to overfitting \cite{gan2021iteratively}. 

The embedded approaches are another types of model-aware methods, which integrate the feature selection process into the training of the downstream learner. In particular, the EXtreme Gradient Boosting (XGB) \cite{xgb} and Random Forest (RF)\cite{rf} are two commonly used embedded feature selection algorithms based on decision trees. Lasso\cite{lasso} utilized L1 regularization for input parameter sparsity. They assess the importance of each feature by evaluating its role and frequency of use at splitting decision points.  Lasso\cite{lasso} utilized L1 regularization for input parameter sparsity. Recently, deep learning has gained widespread attention, and an increasing number of studies have focused on using neural networks for embedded feature selection. E.g. DFS \cite{li2016deep} added a sparse connection layer to the input of a multi-layer perceptron (MLP) to learn nonlinear relationships between features through deep connections. LassoNet \cite{lassonet} employed skip connections and L1 regularization to implement feature ignoring. AEFS \cite{han2018autoencoder} jointly learned a self-representation autoencoder model and the importance weights of each feature. Wojtas et al.\cite{wojtas2020feature} employed a joint training approach with two neural networks and incorporated a random local search process into learning to generate optimal feature subset. {\color{black}STG\cite{stg} introduced random gates to the input layer of neural networks, offering theoretical insights of the feature selection procedure based on approximating of Bernoulli distribution}. COMPFS\cite{imrie2022composite} proposed the group-wise feature selection using neural networks to identify sparse groups with minimal overlap. TSFS \cite{mirzaei2020deep} presented a teacher-student framework based feature selection method, which employed a teacher network to learn the optimal representation of low-dimensional data and utilized a student network for feature selection. {\color{black} CD-LSR\cite{xu2023efficient} transforms the original problem into solving for a discrete selection matrix and utilizes the coordinate descent (CD) method for solving, to achieve sparse learning.}


Different from above model-aware methods, the model-free feature selection is independent of the downstream learning algorithm, and  ranks the features according to their statistical properties such as mutual information, correlation coefficient, variance, etc. As a typical  model-free method, ReliefF \cite{relif} measured the importance of feature by comparing the feature differences of instances within and between classes. The Fisher Score\cite{fisher_score} tried another way to measure the importance of feature by calculating the ratio of between-class mean variance to within-class variance. However, it is limited to classification problems. Some research based on mutual information (MRMR \cite{mrmr}, CIFE\cite{CIFE} and DISR\cite{DISR}) aimed to minimize redundancy within feature subsets while maximizing relevance with the target. Nevertheless,  the calculation of mutual information usually incurs a high computational cost,  and does not inherently support continuous features. {\color{black} MIRFFS}.\cite{hancer2018differential} adopted the differential evolution to address both single-objective and multi-objective problems. Inf-FS \cite{ifs} treated potential feature subsets as paths on a graph, based on which features were ranked and screened. {\color{black} ManiFeSt\cite{cohen2023few} proposed a filter feature selection method based on manifold learning and Riemannian geometry.}

\section{METHODOLOGY}
The pipeline of EasyFS is shown in Fig.~\ref{fig:overview}, which contains four basic modules: 1) \textbf{Feature Expansion} through the random projecting network which aims to introduce non-linear combination of original features; 2) \textbf{Correlation Analysis} between the expanded features and downstream tasks based on conditional covariance; 3) \textbf{Redundancy Estimation} of the features based on the code rate of the feature matrix; 4)\textbf{Multi-objective Feature Selection} which aims to select most correlated and less redundant features. The detail of the design will be presented  after the formal problem definition in the following sections.

\subsection{Problem Formulation}
The procedure of feature selection can be formalized as follows. Given a supervised dataset with the input $X \in \mathbb{R}^{d\times n}$,  $d$ is the number of features and $n$ is the number of samples. $x_{i,j}$ represents the $i$-th feature of the $j$-th sample, and $X_i$ denotes the $i$-th feature across all samples. For the single objective regression problems, the labels are denoted as $Y \in \mathbb{R}^{n}$, representing continuous variables. For the classification problems, the labels are $Y \in \mathbb{R}^{n \times c}$, representing discrete variables. We use $S \subseteq \{1,2,\dots,d\}$ to represent the indices of the feature subset. The goal of feature selection is to find a feature subset $X_S = \{X_i: i \in S\}$ such that $\mathcal{L}(f_\theta(X_s), Y)$ is minimized as much as possible. Here, $\mathcal{L}$ is the loss function, and $f_\theta(.)$ is the downstream model.


\subsection{Feature Expansion  Via  Random Projecting}
 Features in high-dimensional feature space of real-world systems tend to be non-independent, and learning the non-linear combination of the features may achieve significantly better performance than the classical linear models as verified by the fast developing deep neural networks. Thus evaluating the significance of the non-linear feature combinations can help to discover potentially more useful features. However, in the model-free feature selection framework, since the downstream learning model is unknown, we have no idea about how the features will be combined in the future. To exploit the potential of possible combinations of features, we propose a lightweight Random Non-linear Projection network, namely RNP for short, to achieve random non-linear combinations of features. 
 
 RNP adopts the randomly connected network like the reservoir model \cite{esn}, which aims to capture the non-linear features from input time-series. As illustrated in Fig.~\ref{fig:overview}, there are three types of nodes in the network: the input nodes, hidden nodes and enhancement nodes. The input nodes are densely connected to the hidden nodes, while the enhancement nodes is sparsely connected to other nodes. The adjacency matrix of the network is $W \in \mathbb{R}^{p \times p}$ where $p$ indicates the size of the network, initialized randomly using the standard normal distribution. The input feature vector $X \in \mathbb{R}^{d}$ is fed into the network, and propagated in the network for multiple rounds to achieve non-linear fusion of the features.  $I_t \in \mathbb{R}^{p}$ denotes the signals transmitted on each node in the $t^{th}$ round of propagation. Initially  $I_0=[X \quad 0]$, where the signals of the first $d$ input nodes are set as the input feature vector $X$, and the ones of left nodes are set as zero. $h_t \in \mathbb{R}^{p}$ represents the internal state of each node in the $t^{th}$ round. $h_0 = 0$.  $M_t \in \mathbb{R}^{p}$ indicates the output signal of each node, which is calculated through the non-linear propagation procedure as follows:   
\begin{equation}
\label{rmn_1}
  I_{t+1} = ReLU(h_t+W I_t)
\end{equation}
\begin{equation}
\label{rmn_2}
  h_{t+1} = h_t+W I_t-I_{t+1}
\end{equation}
\begin{equation}
\label{rmn_3}
  M_{t+1}=M_t+I_{t+1}
\end{equation} 
The ReLU activation function in Eq.~\ref{rmn_1} is used to introduce the non-linear transformation of the signals. The output signals $M_t$ at the final time step, is used as the non-linear expanded features with $p$ dimension, which is defined as $\hat{M}$:
\begin{equation}
\hat{M}= M_T
\end{equation} 
Here $T$ is the number of propagation rounds.

\subsection{Correlation Analysis}
To measure the significance of each expanded feature $\hat{M}$, we calculate the correlation between the features and the downstream  classification/regression tasks. In particular, for the regression problems, Inspired by \cite{chen2017kernel}, we use the conditional covariance to characterize the correlation. The joint distribution of the $i^{th}$ feature $\hat{M_i}$  and the regression target Y can be fitted as a 
a two-dimensional Gaussian distribution as follows:
\begin{equation}
    \label{re_1}
    Y,\hat{M_i}\sim N(\mu_i,\Sigma_i )
\end{equation}
\begin{equation}
    \label{re_2}
\mu_i=\left(\begin{array}{l}\mu_{Y} \\ \mu_{\hat{M_i}}\end{array}\right)
\Sigma_i =
\begin{pmatrix}
    \sigma_{Y\hat{M_i}} &  \sigma_{YY}\phantom{_{iii}}  \\
    \sigma_{\hat{M_i}Y} &  \sigma_{\hat{M_i}\hat{M_i}}
\end{pmatrix}
\end{equation}
The conditional distribution of a two-dimensional Gaussian distribution is also a Gaussian distribution.
\begin{equation}
    \label{re_3}
Y|\hat{M_i}\sim N(\mu_{Y}+\sigma_{Y\hat{M_i}}\sigma_{\hat{M_i}\hat{M_i}}^{-1}(\hat{M_i}-\mu_{\hat{M_i}}),\sigma_{Y|\hat{M_i}})
\end{equation}

The conditional variance $\sigma_{Y|M_i}$ is:
\begin{equation}
    \label{re_4}
\sigma_{Y|M_i}=\sigma_{YY}-\sigma_{Y\hat{M_i}}\sigma_{\hat{M_i}\hat{M_i}}^{-1}\sigma_{\hat{M_i}Y} 
\end{equation}

Smaller $\sigma_{Y|\hat{M_i}}$ indicates higher correlation between $\hat{M_i}$ and $Y$, so the correlation score of the feature $\hat{M_i}$ can be defined as:
\begin{equation}
    L_i = \sigma_{Y|\hat{M_i}} 
\end{equation}


For classification problems, we directly use the standard deviation of the within-class features to measure the correlation between the feature $\hat{M_i}$ and the $k^{th}$ class: 
\begin{equation}
    \label{re_5}
L_i^k= \sqrt{Var(\{\hat{M}_{i,j}| Y_j=k \})}  
\end{equation}
Here $\hat{M}_{i,j}$ indicates the $i^{th}$ expanded feature of the $j^{th}$ instance. Smaller $L_i^k$ indicates the higher correlation between $\hat{M_i}$ and the  $k^{th}$ class.

\subsection{Redundancy Estimation}
One important goal of feature selection is to filter out the redundant features. Mutual information has been broadly used as redundancy measurement by existing feature selection methods \cite{mrmr}\cite{ross2014mutual}. However, the calculation of mutual information for continuous features are timing costly. In this paper, we propose an efficient redundancy metric based on the coding rate (average coding length) of the feature matrix, which is firstly used by $MCR^2$ \cite{mcr2} to measure the compactness of the features.  Given the feature matrix  $\hat{M} \in \mathbb{R}^{p \times n}$, the  coding rate  \cite{mcr2} can be calculated as follows:
\begin{equation}
    \label{redu_1}
R(\boldsymbol{\hat{M}}, \epsilon) = \frac{1}{2} \log \operatorname{det}\left(\boldsymbol{I}+\frac{p}{n \epsilon^{2}} \boldsymbol{\hat{M}}^{\top} \boldsymbol{\hat{M}}\right)
\end{equation}
Here $\epsilon$ is constant to control the magnitude of coding precision. We represent the redundancy of a feature by examining the variation of the coding rate when the feature is removed from the instances. For the  $i^{th}$ feature, the variation of the code rate is:
\begin{equation}
    \label{redu_2}
Q_i = R(\hat{M}, \epsilon)-R(\hat{M}_{\ominus i},\epsilon)
\end{equation}
Here $\hat{M}_{\ominus i}$ indicates the modified feature matrix of $\hat{M}$ by removing the $i^{th}$ row, which is corresponding to the $i^{th}$ feature. Larger $Q_i$ indicates less redundancy of the $i^{th}$ feature. 

The calculation of Eq. \ref{redu_2} has to be run for each feature, so the measurement is costly for high dimensional feature space. To simplify the calculation of, we rewrite the measurement in a compact matrix form in Eq. \ref{redu_3}. 
\begin{equation}
\begin{split}
\label{redu_3}
Q &=\{Q_1,Q_2,...,Q_p\} \\
&=R(\boldsymbol{\hat{M}},\varepsilon) - \frac{1}{2}log(diag(det(G)G^{-1}))\\
\end{split}
\end{equation}
\begin{equation}
\label{redu_4}
G =\boldsymbol{I}+\frac{p-1}{n \epsilon^{2}} \boldsymbol{\hat{M}}^{\top} \boldsymbol{\hat{M}}
\end{equation}
The detailed proof of Eq.~\ref{redu_3} is provided in the {\color{black} Appendix \ref{appendix_redu_proof}}. 

Meanwhile, for the classification problems, the redundancy measurement $Q_i^k$ is calculated for the $k^{th}$  class based on the feature matrix of the instances belonging to the class.



\subsection{Multi-objective Feature Selection }
In order to select most correlated and less redundant features, we combine the correlation measurement $L_i$ and $Q_i$ of the $i^{th}$ feature to evaluate its importance in the regression task:
\begin{equation}
\label{se_3}
S_i=L_i-\lambda_1 Q_i(1-e^{-Q_i})
\end{equation}
Where $\lambda_1$ is a positive constant used to control the fusion ratio. $L_i$ and $Q_i$ are both pre-processed with min-max normalization before fusion. To emphasize non-redundant features, we add a non-linear weight to the redundancy index, which is defined as $1-e^{-Q_i}$. $S_i$  can be used to evaluate the importance of the extended feature $\hat{M}_i$,  where smaller $S_i$ indicates higher correlation and lower redundancy. 

To perform the feature selection on the original features, we still need to obtain the influence of the original features on the extended features and reflect the importance of high-dimensional features back to the original features.  According to  Eq.~\ref{rmn_3}, the extended feature is the accumulation of the non-linear propagation of the original feature. The support of original feature $X_j (j \leq d)$ on the extended feature $\hat{M}_i (i \leq q)$ can be defined based on the propagation matrix $W$ as follows:
\begin{equation}
    \label{he_1}
    E_{ij}= |(W+I)^{T}_{ij}|
\end{equation}
Here $|\cdot|$ indicates the absolute value function.

After sorting $S_i$ for all expanded features in ascending order, we select the top $r\%$ extended features and calculate the importance of the $j^{th}$ original feature as the support of feature on the selected important extended features:
\begin{equation}
\label{se_1}
H_j = {\textstyle{}  \sum_{i \in Top\_r (\{S_i\})}} E_{ij}
\end{equation}
Here $Top\_r(\cdot)$ indicates the index of the top $r\%$ values in $\{S_i|i \leq q\}$. Higher $H_j$ indicates high importance of the feature $X_j$. 

By combining both of the correlation and redundancy measurements, the multi-object feature selection can be done by ranking the original features according to $H_j$ in descendant order. For the classification  tasks, the similar evaluation can be done for feature selection, the detail of which is shown in the {\color{black} Appendix \ref{appendix_class}}. The full algorithm of the EasyFS covering the whole procedure of feature selection on classification/regression is detailed in the {\color{black} Appendix \ref{appendix_algorithm}}.


\begin{table}[htbp]
\normalsize
\centering
\caption{Regression task datasets}
\label{re_dataset}
\begin{tabular}{lrr}
\toprule
Dataset             &  \#Samples   & \#Feature \\
\midrule
COIL-2000       & 5822  & 85   \\
CSD-1000R       & 500   & 100  \\
SLICE           & 53500 & 385  \\
SML             & 4137  & 26   \\
SoyNAM-height   & 5128  & 4236 \\
SoyNAM-oil      & 5128  & 4236 \\
SoyNAM-moisture & 5128  & 4236 \\
SoyNAM-protein  & 5128  & 4236 \\
Wheat599-env1   & 599   & 1279 \\
Wheat599-env2   & 599   & 1279 \\
Wheat599-env3   & 599   & 1279 \\
Wheat599-env4   & 599   & 1279 \\
\bottomrule
\end{tabular}
\end{table}

\section{Experiment}
\subsection{Experimental Configurations}
\textbf{Datasets.} \ We validate our method on 12 regression datasets and 9 classification datasets as shown in Table \ref{re_dataset} and Table \ref{cl_dataset}. Readers can refer to Appendix \ref{appendix_dataset} for detailed description of the datasets.

\textbf{Evaluation metrics.}\ For the regression problem, in order to eliminate the diversity of the value range in different datasets, the Normalized Mean Square Error Loss NMSE is used as an evaluation metric like \cite{rustam2019application}:
\begin{equation}
\label{ex_4}
NMSE=\frac{ {\textstyle \sum_{n}^{i=1}|y_i-\hat{y_i}|^2 } }{\sum_{n}^{i=1}|y_i-\hat{y}|^2}
\end{equation}
Here $\hat{y_i}$ is the predicted label, $y_i$ is the true label, and $\hat{y}$ denotes the mean of the true labels. Meanwhile, for the classification tasks, the average classification accuracy (ACC) based on k-fold cross-validation is adopted as the evaluation metric like {\color{black} \cite{imrie2022composite}}.



\textbf{Experimental setup.}\ For each dataset, 5-fold cross-validation is conducted. For regression tasks, the Support Vector Machine Regression (SVR) with an RBF kernel is adopted as the downstream learner. For classification tasks, the Support Vector Machine Classification (SVM) with a linear kernel is used as the downstream learner. The parameters of the learners are set to their default values in Scikit-learn \cite{sklearn}. All experiments are conducted using a CPU running at 2.20 GHz and a Nvidia 1080ti GPU.

\textbf{Baselines.}\ Our proposed EasyFS is compared with several popular feature selection methods, including the model-free and model-aware algorithm as follows.

\begin{table}[htbp]
\normalsize
\centering
\caption{Classification task datasets}
\label{cl_dataset}
\begin{tabular}{lrrr}
\toprule
Dataset        & \#Samples  & \#Feature & \#Classes \\
\midrule
COIL20     & 1440 & 1024 & 20  \\
ORL        & 400  & 1024 & 40  \\
PIE-05     & 3332 & 1024 & 68  \\
TOX    & 171  & 5748 & 4   \\
UMIST      & 575  & 1024 & 20  \\
SVHN       & 99289 & 3072  & 10  \\
warpAR10P & 130  & 2400 & 10  \\
YALE       & 165  & 1024 & 15  \\
YALEB      & 2414 & 1024 & 38  \\
\bottomrule
\end{tabular}
\end{table}

\textbf{1) Model-aware Methods}
\begin{itemize}
\item{spFSR\cite{akman2023k} is a recently proposed method based on Simultaneous Perturbation Stochastic Approximation (SPSA) with Barzilai and Borwein (BB) non-monotone gains.}
\item{The stochastic gate (STG) \cite{stg} introduces a stochastic gating mechanism at the input layer of neural networks to select important features by gradient descent. }
\item{Lassonet \cite{lassonet} integrates the skip connections into ResNet \cite{resnet} and utilizes L1 regularization to achieve feature selection. }
\item{EXtreme Gradient Boosting (XGB) \cite{xgb} and Random Forests (RF) \cite{rf} are commonly used ensemble learning algorithms based on the tree structure and also serve as embedded feature selection methods.}
\item{LASSO \cite{lasso,muthukrishnan2016lasso} is a classic embedded feature selection algorithm that combines regression with regularization to impose weight constraints.}
\end{itemize}

\textbf{2) Model-free Methods}
\begin{itemize}
\item{SPEC \cite{spec} is based on spectral theory and assesses the importance of features by analyzing the spectral properties of the data.}
\item{ReliefF \cite{Relief} evaluates the importance of features by estimating their ability to distinguish neighboring samples. However, it is only applicable to classification problems.}
\item{Fisher Score \cite{fisher_score} is a statistical-based method for classification problems based on the measurement of the  intra-class and inter-class differences.}
\item{MCFS \cite{MCFS} integrates manifold learning and L1-regularized models to identify important features.}
\item{PPMCC algorithm \cite{ppmcc} utilizes the Pearson correlation coefficient to rank the importance of features.}
\item{MI algorithm \cite{MI} uses the mutual information between features and target variables to measure the feature importance.}
\end{itemize}


\vspace{-0.5cm}
\begin{figure}[ht]
\vskip 0.2in
\begin{center}
\centerline{\includegraphics[width=\columnwidth]{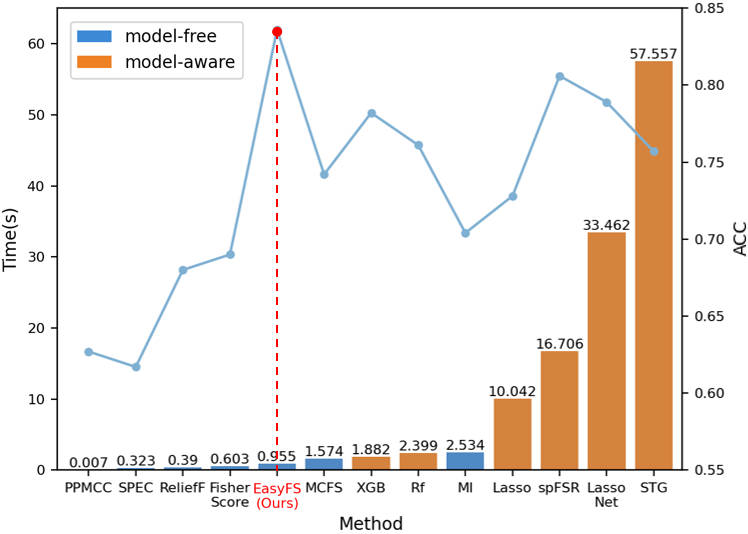}}
\caption{Runtime (by column) and Accuracy (by curve) on the YALE dataset}
\label{method_time}
\end{center}
\vskip -0.2in
\end{figure}

\vspace{-0.4cm}
\begin{figure}[ht]
\vskip 0.2in
\begin{center}
\centerline{\includegraphics[width=0.8\columnwidth]{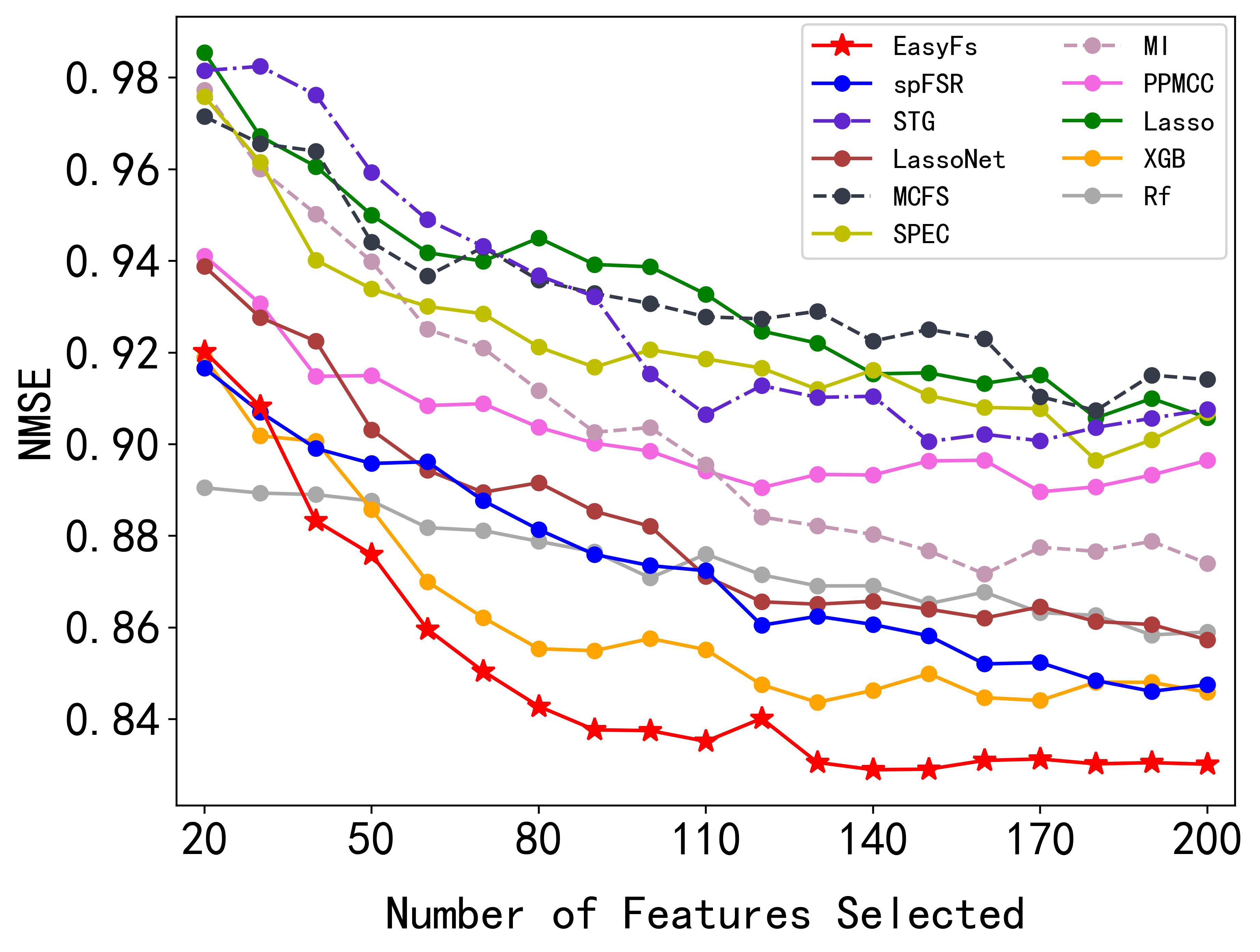}}
\caption{NMSE on the regression dataset SoyNAM-height. (Smaller NMSE is better)}
\label{nmse-SoyNAM-height}
\end{center}
\vskip -0.2in
\end{figure}
\begin{table*}[htbp]\scriptsize
\caption{Regression Results of NMSE (Smaller is better)}
\label{regress_res}
\centering
\renewcommand\arraystretch{1.4}
\begin{threeparttable}[b]
\begin{tabular}{lccccccccccccc}

\toprule
Dataset                          & N\_s & spFSR        & LassoNet & STG          & Rf           & XGB          & Lasso        & PPMCC        & MI           & SPEC   & MCFS   & \begin{tabular}[c]{@{}c@{}}EasyFS\\ (Ours)\end{tabular}           &     \\ \hline
\multirow{2}{*}{COIL-2000}       & 10   & 0.0125       & 0.0145   & 0.7610       & 0.0122       & 0.0121       & {\ul 0.0120} & 0.0127       & 0.0120       & 0.0122 & 0.7645 & \textbf{0.0115} & {\color{red}↓}4.1\%  \\
                                 & 20   & 0.0122       & 0.0144   & 0.7393       & 0.0118       & 0.0118       & {\ul 0.0116} & 0.0121       & 0.0119       & 0.0121 & 0.6812 & \textbf{0.0115} & {\color{red}↓}1.3\%  \\ \hline
\multirow{2}{*}{CSD-1000R}       & 10   & 0.8027       & 0.8251   & 0.8647       & 0.7731       & 0.7818       & {\ul 0.7346} & 0.7790       & 0.8280       & 0.8103 & 0.8184 & \textbf{0.6782} & {\color{red}↓}7.7\%  \\
                                 & 20   & 0.7478       & 0.7611   & 0.7890       & 0.6707       & 0.7482       & {\ul 0.6502} & 0.7284       & 0.7610       & 0.8056 & 0.6738 & \textbf{0.6163} & {\color{red}↓}5.2\%  \\ \hline
\multirow{2}{*}{SLICE}           & 10   & 0.5634       & 0.5335   & 0.6482       & 0.5561       & 0.5664       & {\ul 0.4217} & 0.6906       & 0.8730       & 0.9444 & 0.6110 & \textbf{0.4047} & {\color{red}↓}4.0\%  \\
                                 & 20   & 0.4245       & 0.4659   & 0.5107       & 0.4845       & 0.4216       & {\ul 0.3540} & 0.6866       & 0.8228       & 0.8310 & 0.4974 & \textbf{0.3379} & {\color{red}↓}4.5\%  \\ \hline
\multirow{2}{*}{SML}             & 10   & 0.0635       & 0.0446   & 0.4834       & 0.0414       & 0.0427       & 0.0411       & {\ul 0.0383} & 0.0410       & 0.0592 & 0.1090 & \textbf{0.0373} & {\color{red}↓}2.6\%  \\
                                 & 20   & 0.0365       & 0.0369   & 0.0431       & 0.0353       & {\ul 0.0350} & 0.0350       & 0.0359       & 0.0353       & 0.0390 & 0.0351 & \textbf{0.0350} & {\color{red}↓}0.1\%  \\ \hline
\multirow{2}{*}{SoyNAM-height}   & 50   & 0.8958       & 0.9031   & 0.9593       & 0.8876       & {\ul 0.8858} & 0.9500       & 0.9150       & 0.9399       & 0.9339 & 0.9441 & \textbf{0.8760} & {\color{red}↓}1.1\%  \\
                                 & 150  & 0.8582       & 0.8640   & 0.9006       & 0.8652       & {\ul 0.8499} & 0.9156       & 0.8964       & 0.8768       & 0.9106 & 0.9251 & \textbf{0.8291} & {\color{red}↓}2.5\%  \\ \hline
\multirow{2}{*}{SoyNAM-moisture} & 50   & 0.8893       & 0.9201   & 1.0024       & {\ul 0.8841} & 0.9119       & 0.9956       & 0.8889       & 0.9313       & 0.9757 & 0.9834 & \textbf{0.8790} & {\color{red}↓}0.6\%  \\
                                 & 150  & 0.8788       & 0.9054   & 0.9908       & {\ul 0.8703} & 0.8844       & 0.9744       & 0.8776       & 0.9091       & 0.9448 & 0.9449 & \textbf{0.8508} & {\color{red}↓}2.2\%  \\ \hline
\multirow{2}{*}{SoyNAM-oil}      & 50   & 0.9392       & 0.9057   & 0.9835       & {\ul 0.9039} & 0.9104       & 0.9781       & 0.9225       & 0.9859       & 0.9551 & 0.9508 & \textbf{0.8872} & {\color{red}↓}1.8\%  \\
                                 & 150  & 0.9019       & 0.8897   & 0.9651       & 0.8779       & {\ul 0.8733} & 0.9397       & 0.9174       & 0.9495       & 0.9301 & 0.9450 & \textbf{0.8694} & {\color{red}↓}0.5\%  \\ \hline
\multirow{2}{*}{SoyNAM-protein}  & 50   & 0.8914       & 0.9586   & 0.9608       & {\ul 0.8836} & 0.8906       & 0.9467       & 0.9122       & 0.9467       & 0.9388 & 0.9379 & \textbf{0.8754} & {\color{red}↓}0.9\%  \\
                                 & 150  & 0.8771       & 0.9166   & 0.9715       & 0.8820       & {\ul 0.8692} & 0.9179       & 0.9234       & 0.9266       & 0.9050 & 0.9277 & \textbf{0.8523} & {\color{red}↓}1.9\%  \\ \hline
\multirow{2}{*}{wheat599-env1}   & 50   & 0.8135       & 0.8225   & {\ul 0.7653} & 0.8506       & 0.8062       & 0.8129       & 0.8378       & 0.9054       & 0.9072 & 0.8000 & \textbf{0.7363} & {\color{red}↓}3.8\%  \\
                                 & 150  & 0.7568       & 0.7604   & 0.7642       & 0.7779       & {\ul 0.7514} & 0.7619       & 0.7586       & 0.7766       & 0.8254 & 0.7590 & \textbf{0.6694} & {\color{red}↓}10.9\% \\ \hline
\multirow{2}{*}{wheat599-env2}   & 50   & {\ul 0.8983} & 0.9417   & 0.9094       & 0.9285       & 0.9236       & 1.0162       & 0.9806       & 0.9192       & 0.9641 & 1.0003 & \textbf{0.8462} & {\color{red}↓}5.8\%  \\
                                 & 150  & {\ul 0.8235} & 0.8461   & 0.8719       & 0.8300       & 0.8392       & 0.8739       & 0.8576       & 0.8559       & 0.9923 & 0.8584 & \textbf{0.7941} & {\color{red}↓}3.6\%  \\ \hline
\multirow{2}{*}{wheat599-env3}   & 50   & {\ul 0.9282} & 0.9470   & 0.9415       & 0.9569       & 0.9691       & 0.9914       & 0.9814       & 1.0327       & 1.0037 & 1.0136 & \textbf{0.8842} & {\color{red}↓}4.7\%  \\
                                 & 150  & 0.8912       & 0.9237   & 0.8916       & {\ul 0.8904} & 0.9364       & 0.9233       & 0.9101       & 0.9546       & 0.9919 & 0.9148 & \textbf{0.8552} & {\color{red}↓}3.9\%  \\ \hline
\multirow{2}{*}{wheat599-env4}   & 50   & 0.8739       & 0.8962   & {\ul 0.8245} & 0.8425       & 0.8710       & 0.8578       & 0.9062       & 0.8566       & 0.9823 & 0.9280 & \textbf{0.8076} & {\color{red}↓}2.0\%  \\
                                 & 150  & 0.8238       & 0.8193   & 0.8193       & 0.8407       & 0.8357       & 0.8514       & 0.8487       & {\ul 0.8146} & 0.9742 & 0.8900 & \textbf{0.7646} & {\color{red}↓}6.1\% \\ \bottomrule
\end{tabular}
    \begin{tablenotes}
        \item $N_s$ represents the number of selected features, and the best performance for the subset selection problem is highlighted in bold, while the second-best performance is indicated with an underscore.
    \end{tablenotes}
\end{threeparttable}
\end{table*}

\begin{table*}[htbp]\scriptsize

\centering
\caption{Classification Results of ACC (Larger is better)}
\label{class_res}
\renewcommand\arraystretch{1.4}
\begin{threeparttable}[b]
\begin{tabular}{lccccccccccccccc}
\toprule
Dataset                    & N\_s & spFSR                & \begin{tabular}[c]{@{}c@{}}Lasso\\ Net\end{tabular} & STG   & Rf    & XGB            & Lasso & PPMCC & MI    & ReliefF & SPEC  & MCFS  & \begin{tabular}[c]{@{}c@{}}Fisher\\ Score\end{tabular} & \begin{tabular}[c]{@{}c@{}}EasyFS\\ (Ours)\end{tabular} & \multicolumn{1}{c}{ } \\ \hline
\multirow{2}{*}{COIL20}    & 30   & 0.916                & 0.876                                               & 0.790 & 0.829 & {\ul 0.924}    & 0.849 & 0.598 & 0.756 & 0.744   & 0.664 & 0.827 & 0.741                                                  & \textbf{0.944}                                         & {\color{red}↑}2.2\%                   \\
                           & 50   & 0.951                & 0.942                                               & 0.872 & 0.910 & {\ul 0.964}    & 0.915 & 0.697 & 0.824 & 0.815   & 0.775 & 0.912 & 0.835                                                  & \textbf{0.981}                                         & {\color{red}↑}1.8\%                   \\ \hline
\multirow{2}{*}{ORL}       & 30   & {\ul 0.882}          & 0.823                                               & 0.795 & 0.812 & 0.823          & 0.815 & 0.615 & 0.642 & 0.688   & 0.557 & 0.805 & 0.765                                                  & \textbf{0.893}                                         & {\color{red}↑}1.2\%                   \\
                           & 50   & {\ul 0.925}          & 0.893                                               & 0.893 & 0.905 & 0.907          & 0.887 & 0.705 & 0.735 & 0.767   & 0.725 & 0.878 & 0.838                                                  & \textbf{0.950}                                         & {\color{red}↑}2.7\%                   \\ \hline
\multirow{2}{*}{PIE-05}       & 30   & { \textbf{0.944}} & 0.905                                               & 0.943 & 0.824 & 0.901          & 0.914 & 0.857 & 0.787 & 0.828   & 0.750 & 0.881 & 0.637                                                  & {\textbf{0.944}}                                   & {\color{red}↑}0.0\%                   \\
                           & 50   & {\ul 0.970}          & 0.952                                               & 0.969 & 0.903 & 0.959          & 0.964 & 0.925 & 0.851 & 0.902   & 0.899 & 0.947 & 0.774                                                  & \textbf{0.977}                                         & {\color{red}↑}0.7\%                   \\ \hline
\multirow{2}{*}{TOX}       & 30   & 0.748                & {\ul 0.830}                                         & 0.656 & 0.714 & 0.743          & 0.731 & 0.690 & 0.749 & 0.649   & 0.579 & 0.696 & 0.626                                                  & \textbf{0.860}                                         & {\color{red}↑}3.6\%                   \\
                           & 50   & 0.778                & {\ul 0.860}                                         & 0.778 & 0.784 & 0.789          & 0.696 & 0.749 & 0.772 & 0.678   & 0.708 & 0.732 & 0.684                                                  & \textbf{0.895}                                         & {\color{red}↑}4.1\%                   \\ \hline
\multirow{2}{*}{UMIST}     & 30   & 0.962                & 0.941                                               & 0.934 & 0.927 & \textbf{0.972} & 0.937 & 0.586 & 0.788 & 0.816   & 0.798 & 0.901 & 0.917                                                  & {\ul 0.963}                                            & {\color{green}↓}0.9\%                   \\
                           & 50   & 0.976                & 0.962                                               & 0.950 & 0.965 & {\ul 0.977}    & 0.970 & 0.732 & 0.857 & 0.894   & 0.890 & 0.970 & 0.958                                                  & \textbf{0.979}                                         & {\color{red}↑}0.2\%                   \\ \hline
\multirow{2}{*}{warpAR10P} & 30   & 0.892                & 0.869                                               & 0.823 & 0.877 & {\ul 0.908}    & 0.815 & 0.638 & 0.815 & 0.800   & 0.562 & 0.831 & 0.862                                                  & \textbf{0.946}                                         & {\color{red}↑}4.2\%                   \\
                           & 50   & 0.946                & 0.908                                               & 0.846 & 0.900 & {\ul 0.962}    & 0.823 & 0.769 & 0.862 & 0.815   & 0.569 & 0.908 & 0.923                                                  & \textbf{0.985}                                         & {\color{red}↑}2.4\%                   \\ \hline
\multirow{2}{*}{YALE}      & 30   & {\ul 0.675}          & 0.636                                               & 0.600 & 0.588 & 0.600          & 0.430 & 0.358 & 0.576 & 0.436   & 0.352 & 0.576 & 0.594                                                  & \textbf{0.697}                                         & {\color{red}↑}3.3\%                   \\
                           & 50   & {\ul 0.775}          & 0.715                                               & 0.648 & 0.667 & 0.709          & 0.564 & 0.418 & 0.624 & 0.558   & 0.418 & 0.667 & 0.576                                                  & \textbf{0.818}                                         & {\color{red}↑}5.5\%                   \\ \hline
\multirow{2}{*}{YALEB}     & 30   & {\ul 0.833}          & 0.788                                               & 0.826 & 0.797 & 0.687          & 0.623 & 0.718 & 0.764 & 0.643   & 0.664 & 0.628 & 0.579                                                  & \textbf{0.850}                                         & {\color{red}↑}2.0\%                   \\
                           & 50   & {\ul 0.913}          & 0.889                                               & 0.911 & 0.901 & 0.847          & 0.766 & 0.834 & 0.876 & 0.805   & 0.814 & 0.806 & 0.708                                                  & \textbf{0.920}                                         & {\color{red}↑}0.8\%                   \\ \hline
\multirow{2}{*}{SVHN}      & 30   & {\ul 0.204}          & 0.200                                               & 0.194 & 0.198 & 0.201          & 0.200 & 0.196 & 0.200 & 0.199   & 0.194 & 0.193 & 0.198                                                  & \textbf{0.215}                                         & {\color{red}↑}5.4\%                   \\
                           & 50   & {\ul 0.212}          & 0.206                                               & 0.193 & 0.201 & 0.205          & 0.203 & 0.196 & 0.202 & 0.197   & 0.193 & 0.194 & 0.200                                                  & \textbf{0.224}                                         & {\color{red}↑}5.7\%                   \\ \bottomrule
\end{tabular}
    \begin{tablenotes}
        \item $N_s$ represents the number of selected features, and the best performance for the subset selection problem is highlighted in bold, while the second-best performance is indicated with an underscore.
    \end{tablenotes}
\end{threeparttable}
\end{table*}
\subsection{Experimental Result}
\begin{figure}[ht]
\vskip 0.2in
\begin{center}
\centerline{\includegraphics[width=0.8\columnwidth]{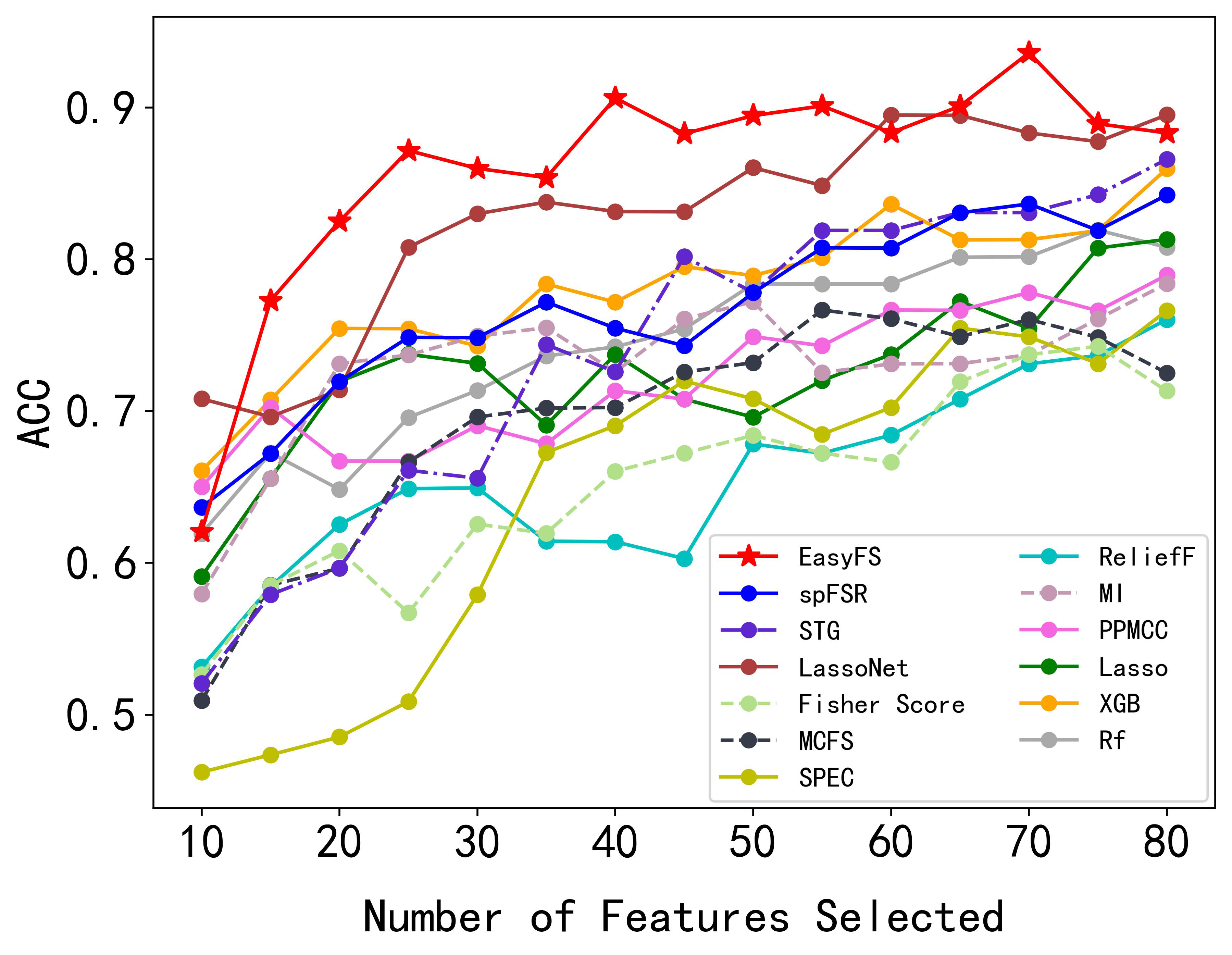}}
\caption{ACC on the classification dataset Tox. (Higher ACC is better)}
\label{acc-tox}
\end{center}
\end{figure}

We compare the performance of various feature selection methods on both regression and classification tasks in Tables \ref{class_res} and \ref{regress_res}, where $N_s$ represents the number of selected feature subsets. We also depict comparison curves for selected datasets at each step of two typical datasets will highest dimensions in Fig. \ref{acc-tox} and Fig. \ref{nmse-SoyNAM-height}. Reader can refer to {\color{black} Appendix \ref{appendix_result_pic}} for the full results on all datasets. The results indicate that our proposed EasyFS outperforms the aforementioned feature selection methods in almost all cases. In particular, the superiority of EasyFS  over the the second-best method by 5.7\% on the SVHN classification dataset, and 10.9\%  On the wheat599-env1 regression dataset. For other feature selection algorithms, spFSR, STG, and XGB exhibit strong performance in classification tasks, while Lasso and spFSR perform well in regression tasks. Furthermore, we adding the visualization analysis of all methods to verify the effectivess of the model in the {\color{black} Appendix \ref{appendix_vis}}.
\vspace{-0.5cm}
\vspace{-0.3cm}
Additionally, we compared the runtime required by different algorithms to select the top 5\% of features during feature selection on the YALE dataset in Fig.~\ref{method_time}. It can be observed that the model-free algorithms such as ReliefF, PPMCC, and Fisher Score exhibit the fastest execution times, but have relatively much worse performance of the downstream classification task. In contrast, algorithms based on gradient descent, such as STG, LassoNet, and Lasso, have better classification performance, but require much more time, e.g. STG even taking 146 times longer than the ReliefF algorithm. Our method EasyFS have the highest classification performance while saving more than 94\% time compared to the latest model-aware method spFSR.


\textbf{Ablation.}\ Based on the Wheat599-env1 dataset, we conducted experiments to test the effectiveness of three important modules of DeepFS: Random Non-linear Projection (RNP), Redundancy Estimation (RE), and Correlation Analysis (CA). As shown in  Table \ref{Ablation}, adding the RNP structure reduces the error by 10.87\%, while building the RNP and its propagation process is also the main part of algorithm time consumption. The addition of CR reduced the error by 3.1\% with a relatively minor impact on the running speed.
\begin{table}[]
\normalsize
\centering
\renewcommand\arraystretch{1.3}
\caption{Ablation Experiment}
\label{Ablation}
\begin{tabular}{c|ccccc}
\toprule
\multicolumn{1}{l|}{Dataset}                                              & RNP & CR & CA & Time(s) & NMSE       \\ \hline
\multirow{4}{*}{\begin{tabular}[c]{@{}c@{}}Wheat599\\ -env1\end{tabular}} & $\surd$   & $\surd$  & $\surd$  & 1.324   & \textbf{0.713} \\
                                                                          & $\surd$   & $\times$  &$\surd$  & 1.031   & 0.746          \\
                                                                          & $\times$   & $\surd$  & $\surd$  & 0.128   & 0.811          \\
                                                                          & $\times$   & $\times$  &$\surd$  & 0.007   & 0.837          \\ \bottomrule
\end{tabular}
\end{table}

\begin{figure}[ht]
\vskip 0.2in
\begin{center}
\centerline{\includegraphics[width=\columnwidth]{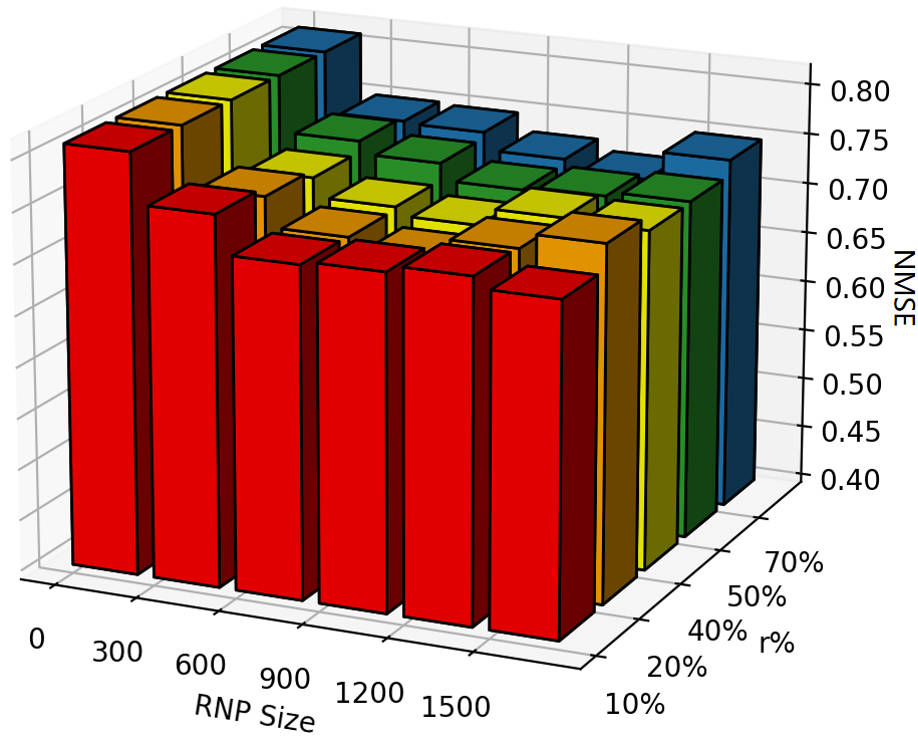}}
\caption{Parameter Analysis}
\label{para_analysis}
\end{center}
\vskip -0.2in
\end{figure}

\textbf{Parameter Analysis.}\ In EasyFS, the size of the Random Non-linear Projecting Network (RNP) and the hyperparameter $r$ for selecting the top $r\%$ high-dimensional features can influence the experimental results. We varied the feature range expanded by RNP from $\{0,300,...,1500\}$ and $r$ from $\{0.1,0.2,...,0.7\}$, and the experimental results are shown in Fig. \ref{para_analysis}. It can be observed that as the RNP size increases, the error rapidly decreases and then stabilizes with small fluctuations. The choice of $r$ needs to be within a reasonable range because selecting too few or too many high-dimensional features can lead to an increase of error. Selecting too many features introduces more unrelated features, while selecting too few may not utilize enough information of non-linear combinations of features.
\vspace{-0.5cm}
\vspace{-0.5cm}

\section{conclusion}
In this paper, we propose a high-performance  model-free feature selection method via elastic transformation of features, namely EasyFS, to achieve the goals of effectiveness, efficiency and flexibility. EasyFS utilizes a Random Non-linear Projection Network  to expanding the feature space and calculate the correlation of high-dimensional features. Meanwhile, a novel redundancy measurement based on coding rate is proposed to decrease the feature dimension efficiently. Comprehensive experiments on both classification and regression tasks, which are conducted based on 21 real-world datasets, show the superior performance of EasyFS on both of the precision of downstream tasks and the timing cost compared with state-of-the-art methods.

In the future, we will extend EasyFS to further support the feature selection in the multi-modal use cases.

\section{Acknowledgements}
This work was supported by National Key R\&D Program
of China (2023YFA1011601, 2023YFA1011602), the Basic and Applied
Basic Research Foundation of Guangdong Province
(2024A1515012287), Science and Technology Key Program of Guangzhou (2023B03J1388)

\bibliography{example_paper}
\bibliographystyle{icml2024}

\newpage
\appendix
\onecolumn
\section{Coding Redundancy Matrix operation formula derivation}
\label{appendix_redu_proof}
The spatial encoding rate formula for a certain batch of samples.
\begin{equation}
\label{ap_1}
R(\boldsymbol{M}, \epsilon)=\frac{1}{2} \log \operatorname{det}\left(\boldsymbol{I}+\frac{p}{n \epsilon^{2}} \boldsymbol{M}^{\top} \boldsymbol{M}\right)\in R^{p \times p}
\end{equation}
If one column (a one-dimensional feature) is removed from M, get  $M_{i'}=\{M_j\mid j\in \{0,1,\ldots,l\},j\ne i\}$, it can be obtained that
\begin{equation}
\label{ap_2}
R(\boldsymbol{M_{i'}}, \epsilon)=\frac{1}{2} \log \operatorname{det}\left(\boldsymbol{I}+\frac{p}{n \epsilon^{2}} \boldsymbol{M_{i'}}^{\top} \boldsymbol{M_{i'}}\right)\in R^{(p-1) \times (p-1)}
\end{equation}
At this point, let
\begin{equation}
\label{ap_3}
R(\boldsymbol{M_{i'}}, \epsilon)=\frac{1}{2}log det(G_{i'})
\end{equation}
\begin{equation}
\label{ap_4}
G=\boldsymbol{I}+\frac{p-1}{n \epsilon^{2}} \boldsymbol{M}^{\top} \boldsymbol{M}\in R^{p \times p}
\end{equation}
Then, $G_{i'}$ is exactly the value of $G$ after removing the i-th row and the i-th column.
\begin{equation}
\label{ap_5}
G=\boldsymbol{I}+\frac{p-1}{n \epsilon^{2}} \boldsymbol{M}^{\top} \boldsymbol{M}\in R^{p \times p}
\end{equation}
We want to calculate the determinant of $G_{i'}$ for each iteration over i, which is ${det(G_{1'}), det(G_{2'}), ..., det(G_{p'})}$, and express it in matrix form.
\begin{equation}
\label{ap_6}
\begin{bmatrix}
    det(G_{1'}) & 0 & \dots & 0 \\
    0 & det(G_{1'}) & \dots & 0 \\
    \vdots & \vdots & \ddots & \vdots \\
    0 & 0 & \dots & det(G_{p'}) \\
\end{bmatrix}
\end{equation}
Then, according to the definition of the adjugate matrix, these terms exactly correspond to the diagonal of the adjugate matrix of G, which is $diag(adj(G))$. The adjugate matrix can be calculated through the inverse of the matrix and its determinant as $adj(G) = det(G)G^{-1}$.

so
\begin{equation}
\label{ap_7}
\{det(G_{1'}),det(G_{2'}),...,det(G_{p'})\} = diag(det(G)G^{-1})
\end{equation}
\begin{equation}
\label{ap_8}
\{R(\boldsymbol{M_{1'}}, \epsilon),R(\boldsymbol{M_{2'}}, \epsilon),...,R(\boldsymbol{M_{n'}}, \epsilon)\}=\frac{1}{2}log(diag(det(G)G^{-1}))
\end{equation}
\begin{equation}
\label{ap_9}
\begin{split}
    Q &= \{Q_1,Q_2,....,Q_n\} \\
     &= R(M,\epsilon)-\{R(\boldsymbol{M_{1'}}, \epsilon),R(\boldsymbol{M_{2'}}, \epsilon),...,R(\boldsymbol{M_{n'}}, \epsilon)\} \\
     &= R(M,\epsilon) - \frac{1}{2}log(diag(det(G)G^{-1}))
\end{split}
\end{equation}


\section{Importance Evaluation of Features in the Classification Tasks}
\label{appendix_class}
For the classification tasks, the importance of the $i^{th}$ extended features for the $k^{th}$ class can be defined similar with Eq.~\ref{se_3}:
\begin{equation}
\label{se_4}
S_i^k=L_i^k-\lambda_1 Q_i^k(1-e^{-Q_i^k})
\end{equation}
Similar with  Eq.~\ref{se_1}, the importance of the $j^{th}$ original feature on the $k^{th}$ class can be defined as:
\begin{equation}
\label{se_5}
H_j^k = {\textstyle{}  \sum_{i \in Top\_r (\{S_i^k\})}} E_{ij}
\end{equation}

Inspired by  Fisher Score\cite{fisher_score}, Laplacian Scoree\cite{lp_score}, and ReliefF\cite{relif}, we further extend Eq.~\ref{se_5} by emphasize features with large inter-class variation as follows:
\begin{equation}
\label{se_8}
H_i^k = {\textstyle{}  \sum_{i \in Top\_r (\{S_i^k\})}} (\sigma_i^\mu+\lambda_2\sigma_i^\sigma) E_{ij}
\end{equation}
Here $\mu_i^k$ and $\sigma_i^k$ represent the mean and variance of the $i$-th feature within class $k$. $\sigma _i^\mu $  represent the variance of the inter-class means and $\sigma _i^\sigma $ to represent the variance of inter-class variances for the i-th feature. 

The importance of the $j^{th}$ original feature is calculated as follows in the end:
\begin{equation}
\label{se_9}
H_i = \sum_{k=1}^cH_i^k
\end{equation}

\section {Full Algorithm of EasyFS}%
\label{appendix_algorithm}
\renewcommand{\algorithmicrequire}{ \textbf{Input:}}
\begin{algorithm}[htbp]
\caption{EasyFS Algorithm.}\label{alg:alg1}
\begin{algorithmic}[1]
\REQUIRE dataset$\{X,Y\}$; Hyperparameter $\lambda _1,\lambda _2,r$; $W$,$T$ for RNP parameter
\STATE Calculate $E_{ij}$ by Eq. (\ref{he_1})
\STATE Calculate $M$,Mapping features $f:\mathbb{R}^d\rightarrow \mathbb{R}^p$ via RNP
\IF{{Regression Task}}
\STATE Calculate $G=\boldsymbol{I}+\frac{p-1}{n \epsilon^{2}} \boldsymbol{M}^{\top}\boldsymbol{M}$
\STATE Calculate $Q=R(M,\varepsilon)-\frac{1}{2}log(diag(det(G)G^{-1}))$
\FOR {i = 1,...,p}
\STATE Calculate correlation $L_i=\sigma_{Y|M_i}=\sigma_{YY}-\sigma_{Y\hat{M_i}}\sigma_{\hat{M_i}\hat{M_i}}^{-1}\sigma_{\hat{M_i}Y} $
\STATE Calculate $S_i=L_i-\lambda_1 Q_i(1-e^{-Q_i})$
\STATE Calculate feature importance $H_i$ by Eq. (\ref{se_1})
\ENDFOR
\ELSIF{{Classification Task}}
\FOR {k = 1,...,c}
\STATE Calculate $G=\boldsymbol{I}+\frac{p-1}{n \epsilon^{2}} \boldsymbol{M^k}^{\top} \boldsymbol{M^k}$
\STATE Calculate $Q^k=R(M^k,\varepsilon)-\frac{1}{2}log(diag(det(G)\allowbreak G^{-1}))$
\FOR {i = 1,...,p}
\STATE Calculate correlation $L_i^k$ by Eq. (\ref{re_5})

\STATE Calculate $S_i^k=L_i^k-\lambda_1 Q_i^k(1-e^{-Q_i^k})$
\STATE Calculate $H_i^k$ by Eq. (\ref{se_8})
\ENDFOR
\ENDFOR
\STATE Calculate feature importance $H_i$ by Eq. (\ref{se_9})
\ENDIF
\STATE Return rank($H_i$)
\end{algorithmic}
\label{alg1}
\end{algorithm}

\section{Dataset Configurations}
\label{appendix_dataset}
\textbf{Regression task datasets.}\ We validate our method on both regression and classification datasets. For regression tasks, we use a total of 12 datasets, including 4 single-target regression datasets and derived subsets from 8 multi-target gene expression datasets. Most of these datasets are sourced from the UCI Machine Learning Repository\cite{uci} and other publicly available datasets. SoyNAM\cite{liu2019phenotype} is a gene dataset comprising the genotype-phenotype relationships of 5128 soybean plants. It uses 4236 genes to predict various phenotypic traits of soybeans, including protein content, oil content, moisture content, and height. Each phenotype is treated as a separate prediction target since different genes influence each phenotype differently. Consequently, the dataset is divided into four distinct sub-datasets: SoyNAM-height, SoyNAM-oil, SoyNAM-moisture, and SoyNAM-protein.
Wheat599\cite{wang2023dnngp} is a dataset consisting of 599 historical wheat lines obtained from the International Maize and Wheat Improvement Center (CIMMYT) Global Wheat Program\cite{mclaren2005international}. The project involved multiple international wheat experiments, categorizing wheat-growing environments into four main climatic regions. The dataset records the yield of the same wheat genotypes in these four different production environments. Similarly, we treat these four environments as four separate targets, resulting in four sub-datasets: Wheat599-env1, Wheat599-env2, Wheat599-env3, and Wheat599-env4.
The COIL-2000\cite{uci} dataset comprises information on 5822 customers of an insurance company. It aims to predict the customer type or class based on individual customer attributes such as whether they have received higher education, the number of children they have, income, and so on.
SLICE\cite{uci} is a medical dataset that contains information from 71 patients and 53,500 CT images, using Histogram describing bone structures and Histogram describing air inclusions to predict Relative location of the image on the axial axis.
The SML dataset\footnote{Data from https://github.com/treforevans/uci\_datasets} collects environmental sensor data from rooftop sensors, including parameters such as temperature, humidity, and wind speed. It comprises 4,371 records of environmental sensor data.
The CSD-1000R dataset, as described in \cite{musil2019fast}, contains atomic environments of C, H, N, and O atoms extracted from the crystal structures of molecular compounds. This dataset is used for predicting their NMR chemical shielding.
The number of samples and the number of features for all regression datasets are shown in Table \ref{re_dataset}.

\begin{figure}[ht]
\vskip 0.2in
\begin{center}
\centerline{\includegraphics[width=0.75\columnwidth]{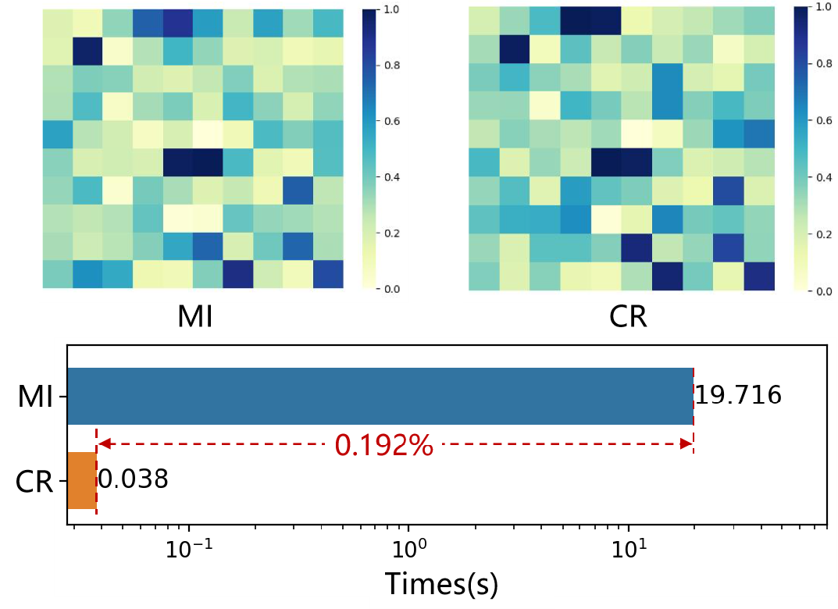}}
\caption{Comparison with mutual information}
\label{compare_to_mi}
\end{center}
\vskip -0.2in
\end{figure}
\textbf{Classification task datasets.}\ For classification tasks, we conducted experiments on 9 datasets.
The COIL20 object recognition dataset\cite{coil20} contains grayscale images of 20 objects captured from different angles. There are 72 images taken for each object, with images taken at 5-degree intervals.
The ORL face dataset\cite{ORL}, consisting of 400 images from 40 different individuals, was collected under various conditions such as different times, lighting conditions, facial expressions (open eyes/closed eyes, smiling/not smiling), and facial details (with glasses/without glasses).
The PIE-05 dataset\cite{PIE} consists of imaging data from 68 volunteers captured under 43 different lighting conditions, with four distinct facial expressions, and across 13 unique poses for each individual.We use the data captured by Camera 05, which comprises a total of 3332 images.
UMIST\cite{UMIST}, warpAR10P\cite{li2017feature}, YALE\cite{YALE}, YALEB\cite{YALEB} are all face recognition datasets.
The TOX dataset\cite{TOX} is a biological dataset that includes 171 samples, each with 5748 genes. It is divided into four categories: non-cancer patients, controlled radiation therapy patients, cancer skin patients, and radiation therapy patients.
SVHN (Street View House Numbers)\cite{svhn} is a digit recognition dataset obtained by cropping images of house numbers from street view scenes. It contains 99,289 color images of digits, each with a resolution of 32x32 pixels. Example of the SVHN dataset as shown in Fig. \ref{svhn_dataset}.The number of samples and features for all classification datasets is shown in Table \ref{cl_dataset}.

\begin{figure*}[htbp]
\centering
\subfloat[Wheat599-env1]{\includegraphics[width=0.32\textwidth]{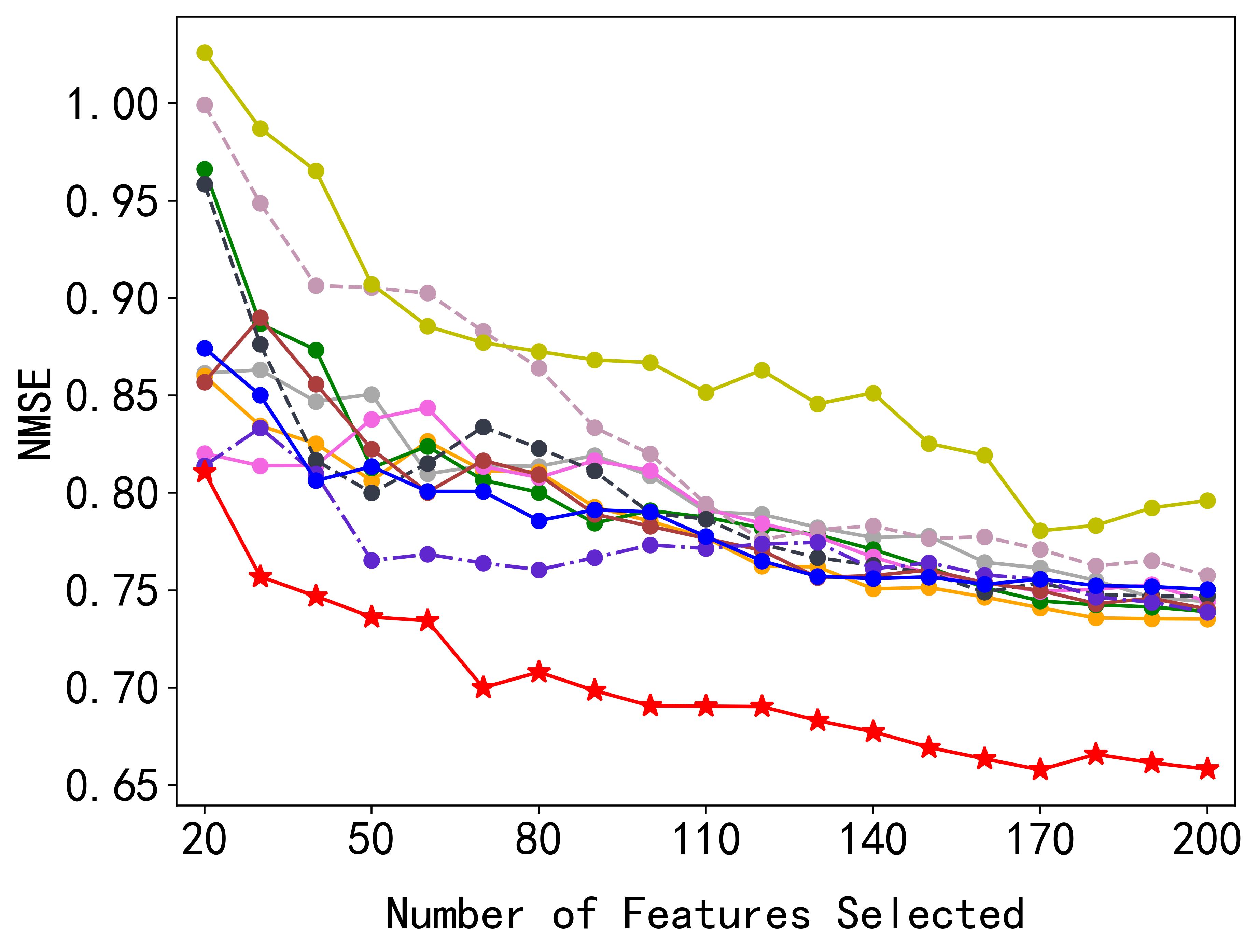}%
\label{fig_second_case}}
\hfil
\subfloat[Wheat599-env2]{\includegraphics[width=0.32\textwidth]{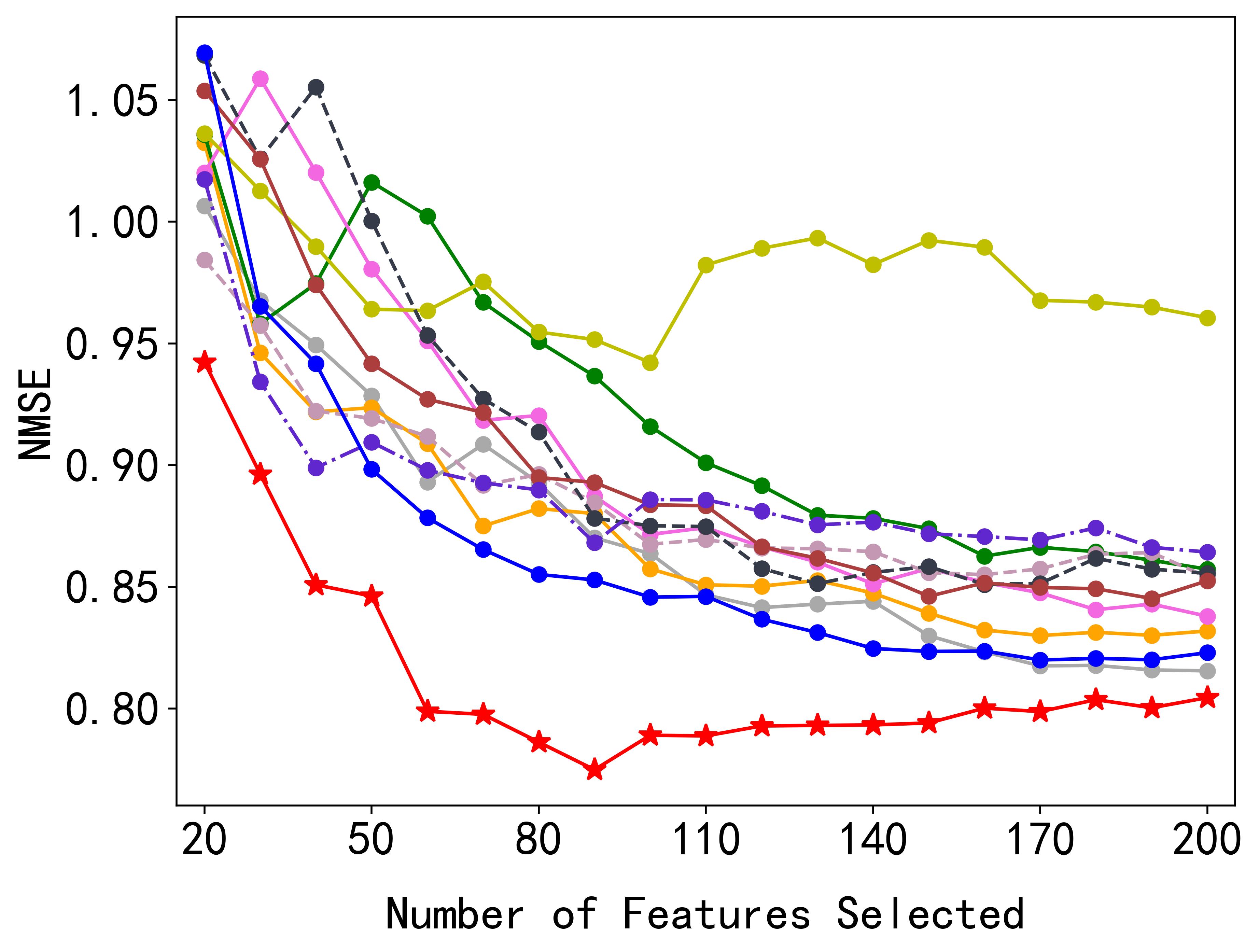}%
\label{fig_second_case}}
\hfil
\subfloat[Wheat599-env3]{\includegraphics[width=0.32\textwidth]{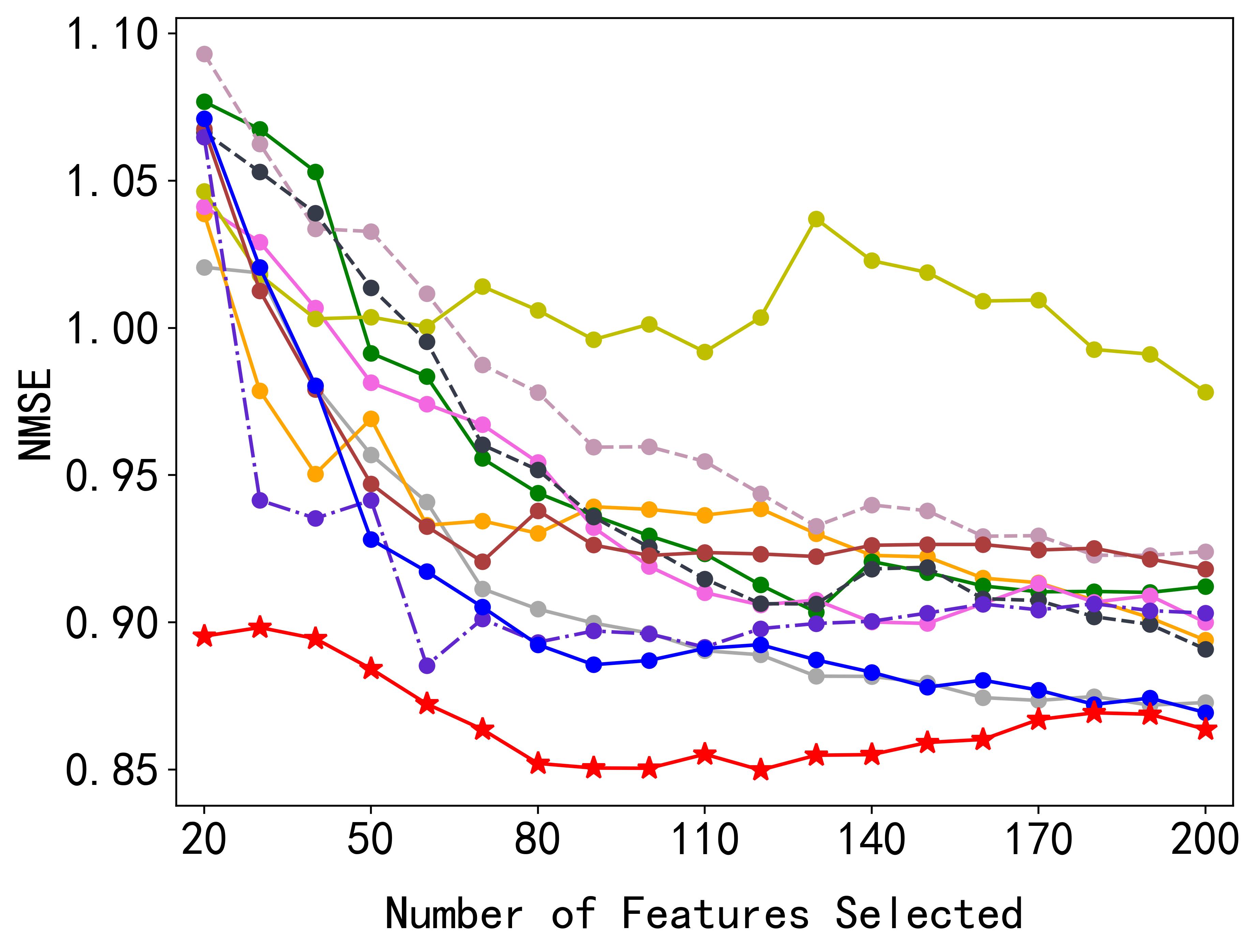}%
\label{fig_second_case}}
\hfil
\subfloat[Wheat599-env4]{\includegraphics[width=0.32\textwidth]{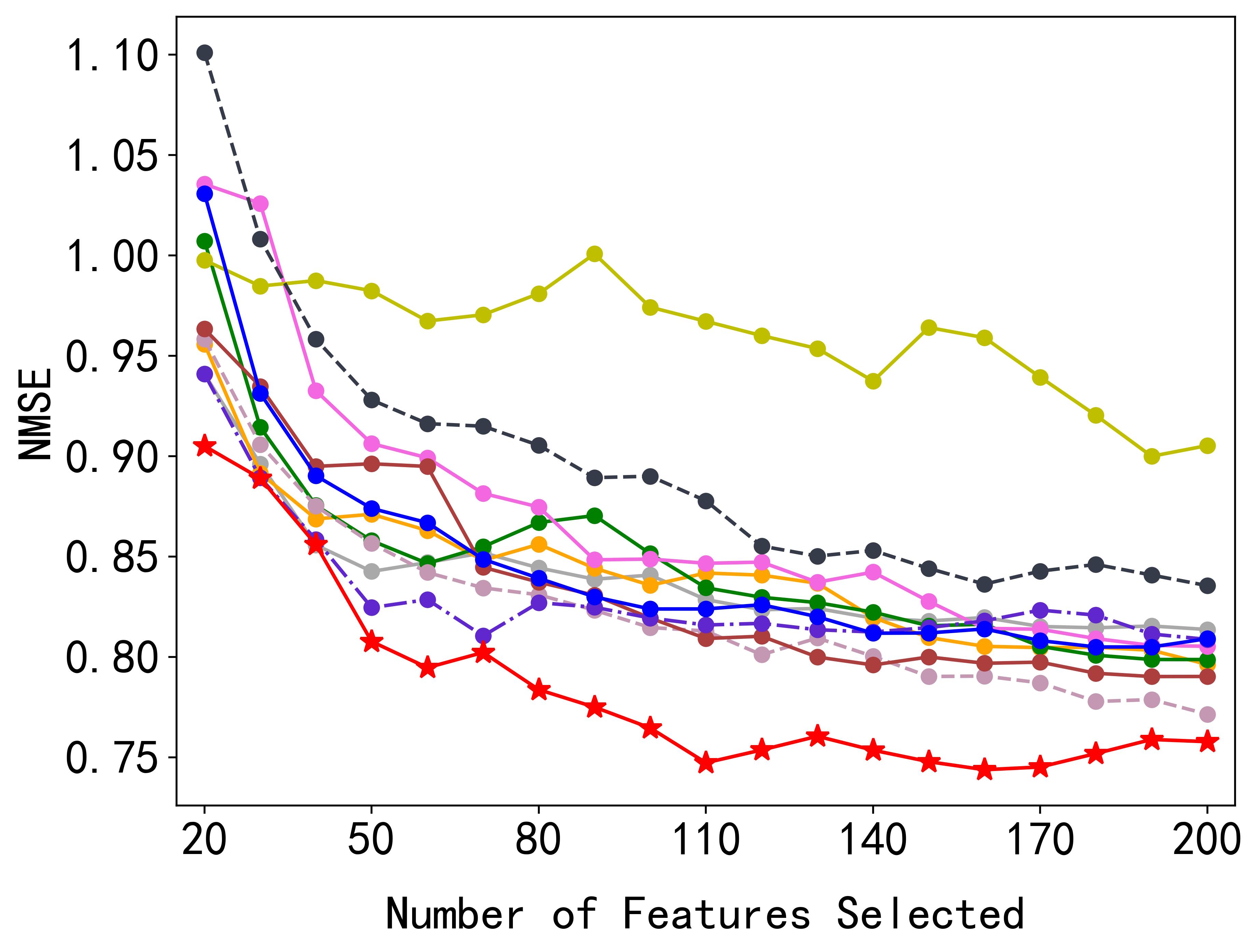}%
\label{fig_second_case}}
\hfil
\subfloat[SoyNAM-height]{\includegraphics[width=0.32\textwidth]{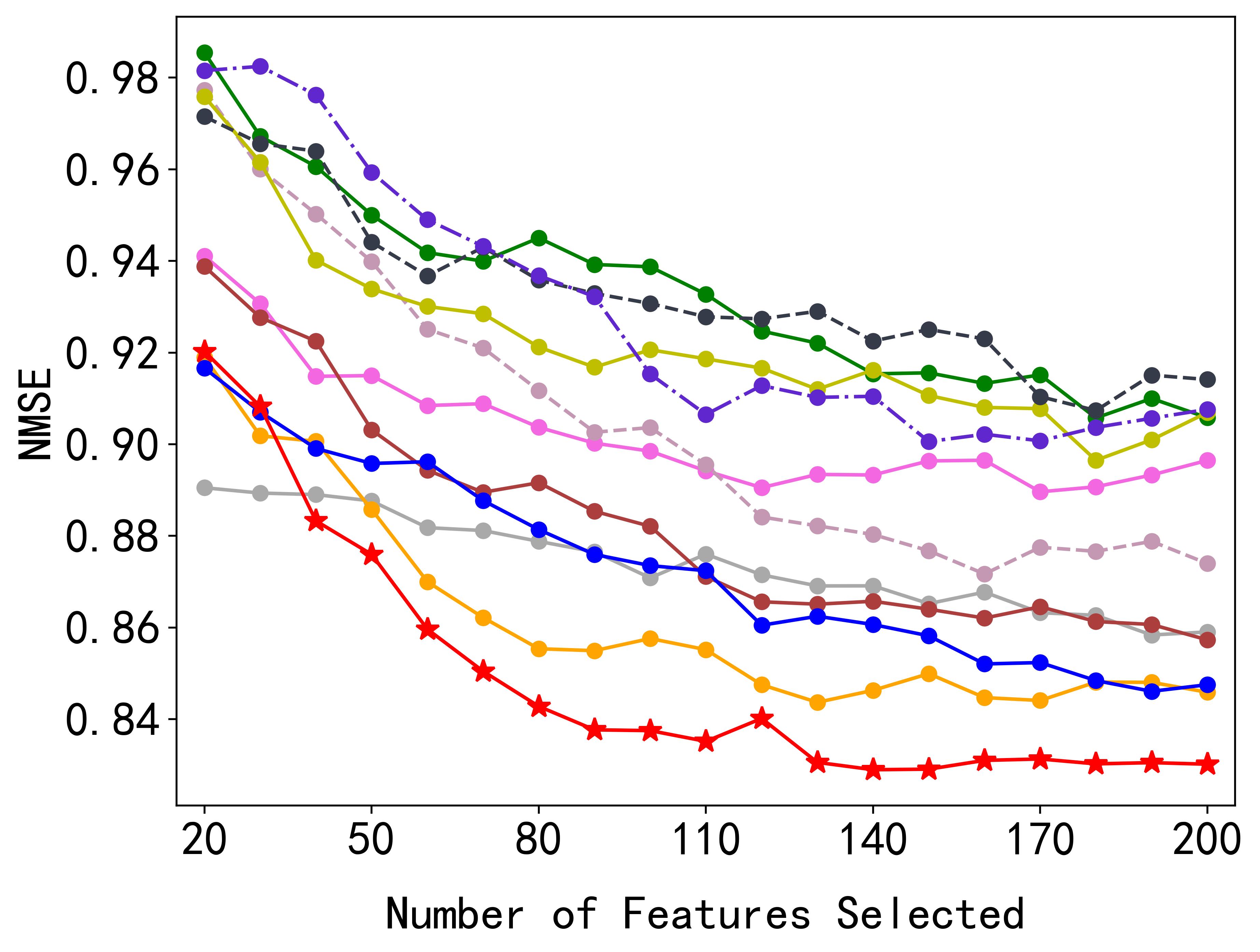}%
\label{fig_second_case}}
\hfil
\subfloat[SoyNAM-oil]{\includegraphics[width=0.32\textwidth]{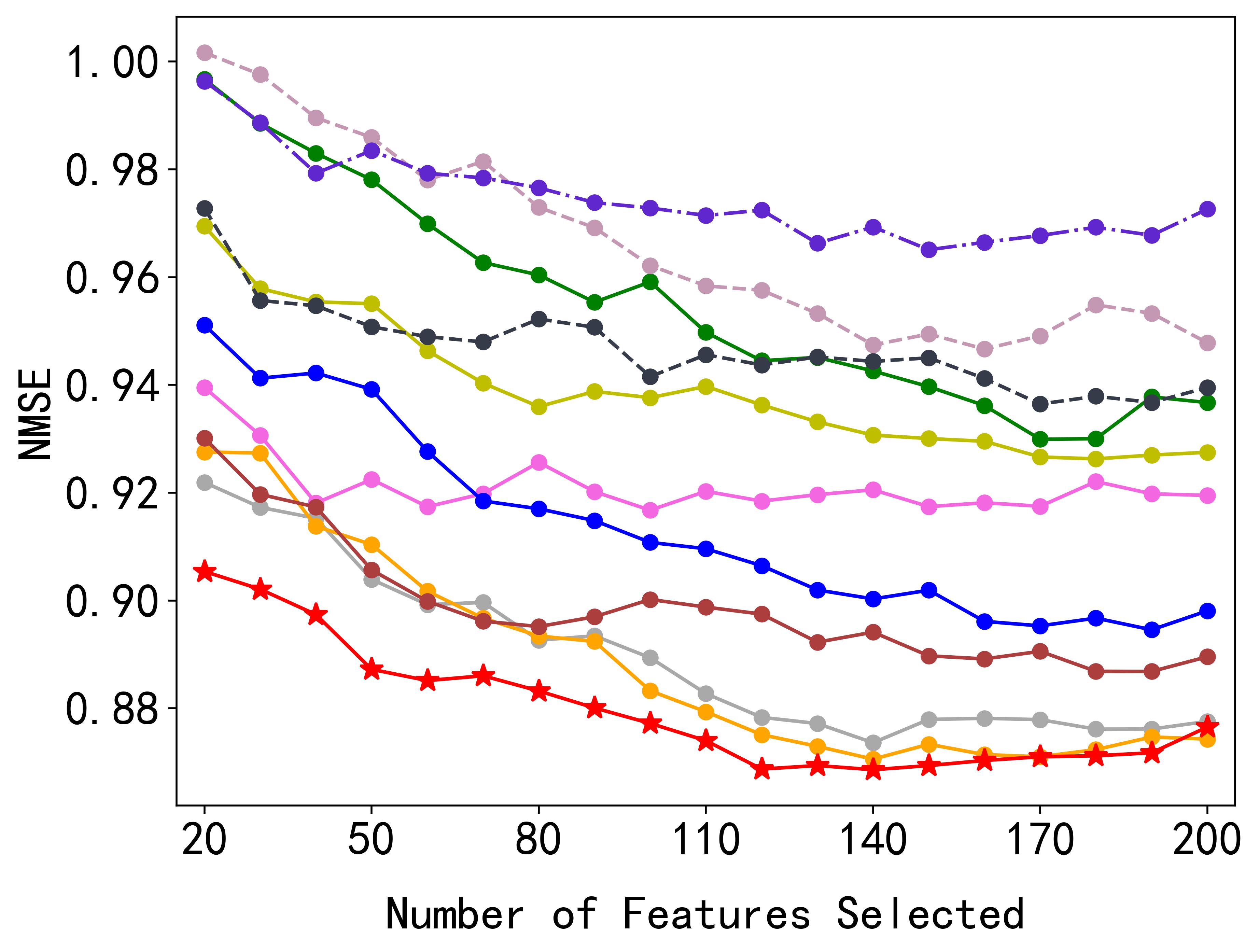}%
\label{fig_second_case}}
\hfil
\subfloat[SoyNAM-moisture]{\includegraphics[width=0.32\textwidth]{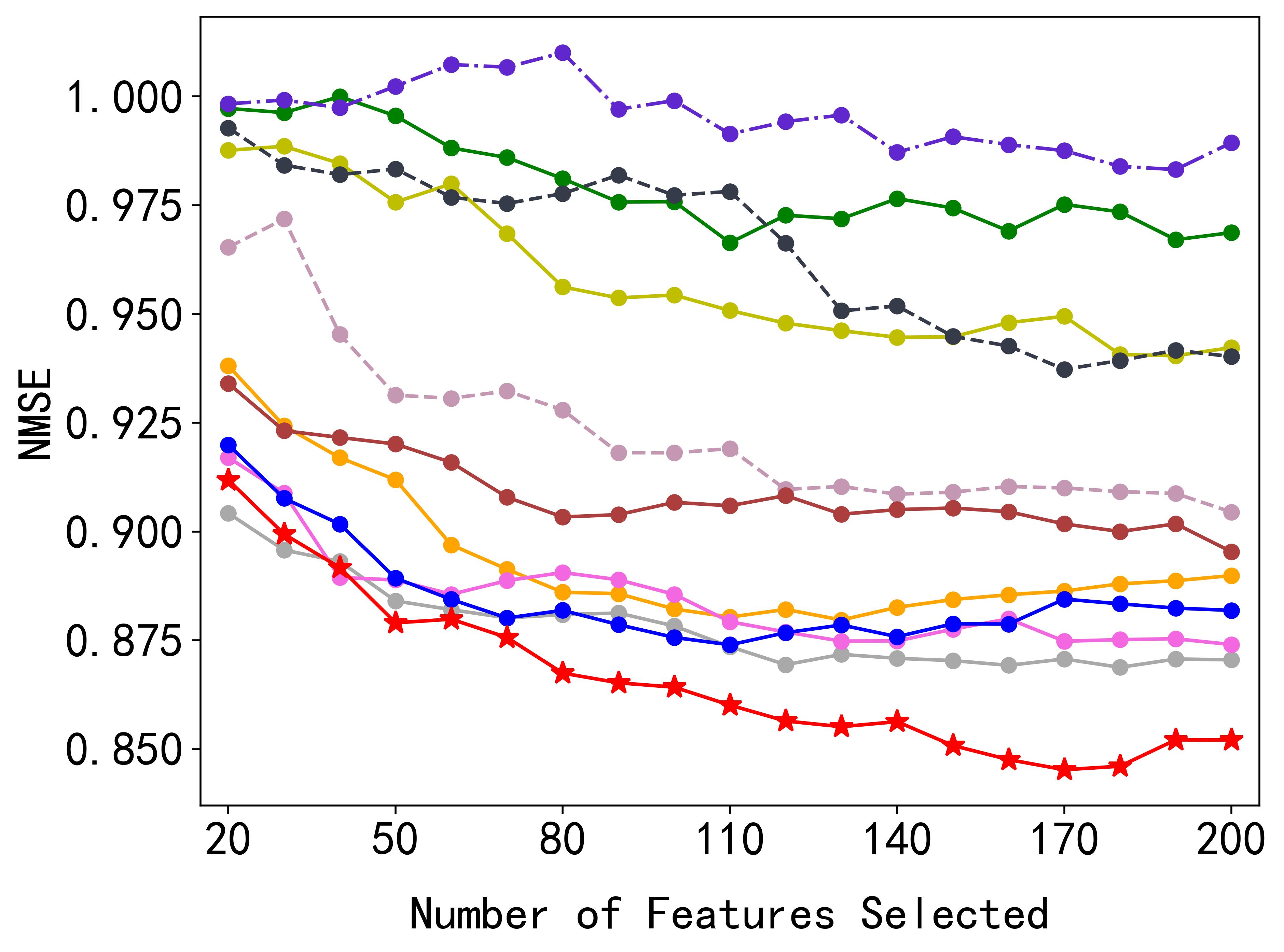}%
\label{fig_second_case}}
\hfil
\subfloat[SoyNAM-protein]{\includegraphics[width=0.33\textwidth]{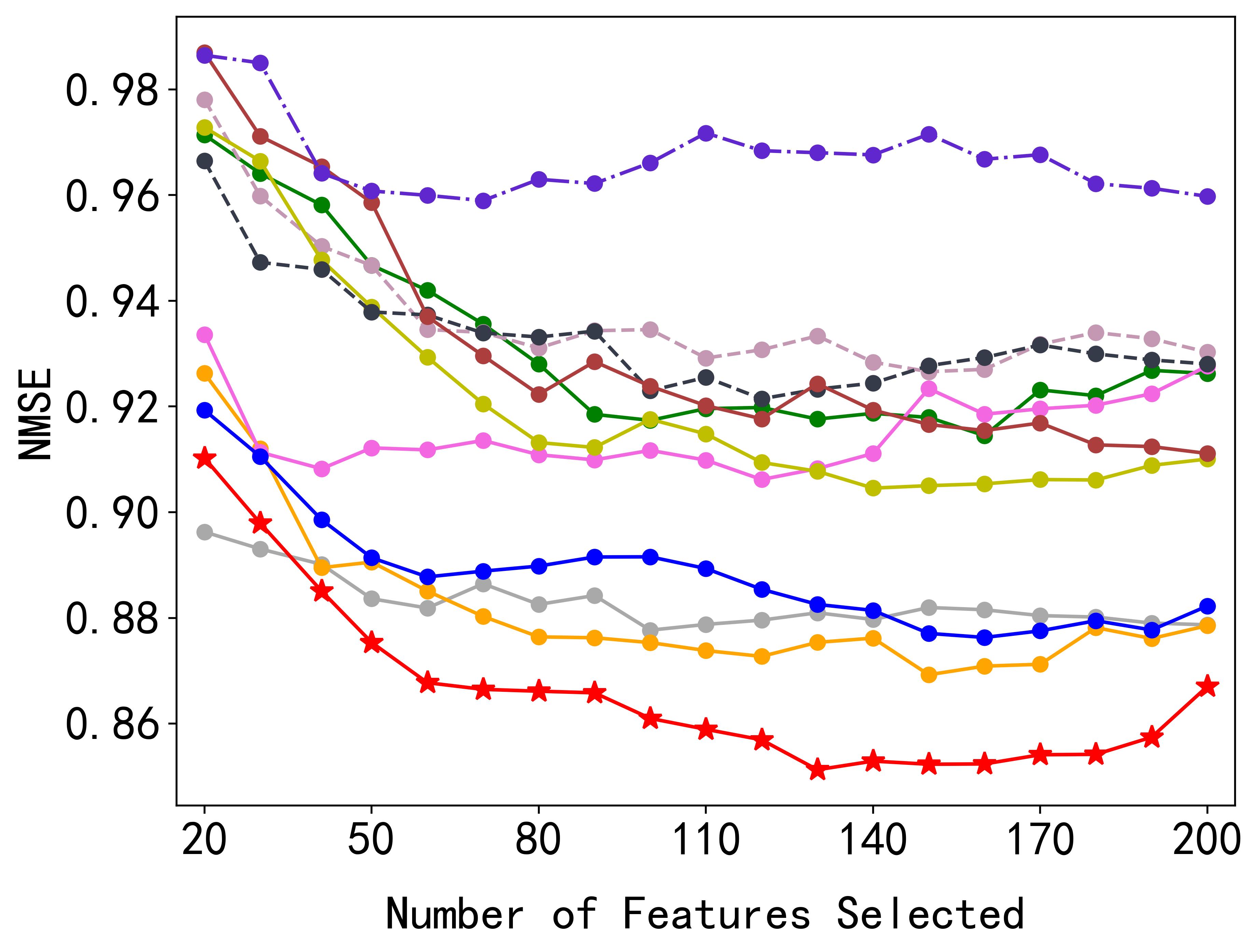}%
\label{fig_second_case}}
\hfil
\subfloat[CSD-1000R]{\includegraphics[width=0.32\textwidth]{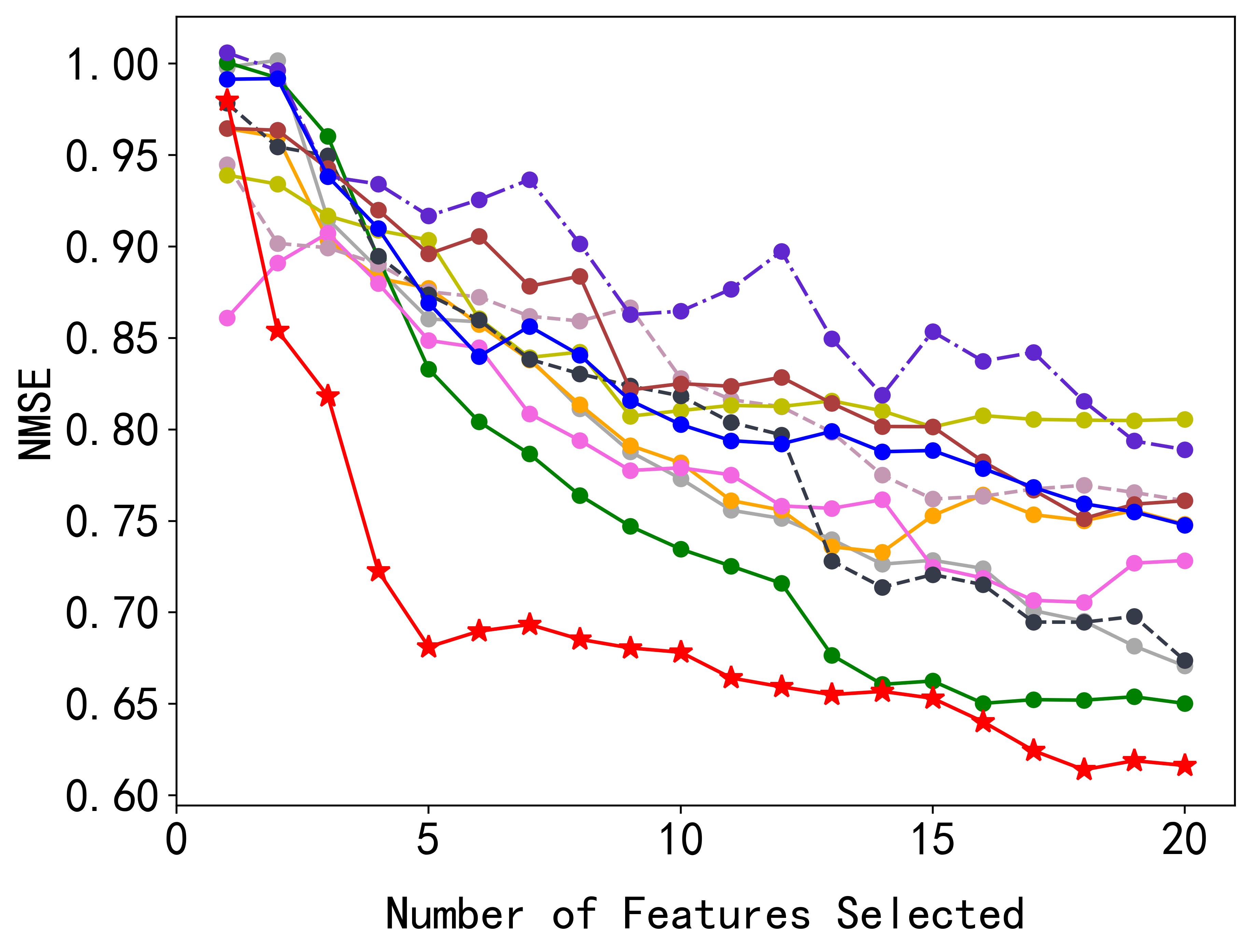}%
\label{fig_first_case}}
\hfil
\subfloat[SLICE]{\includegraphics[width=0.32\textwidth]{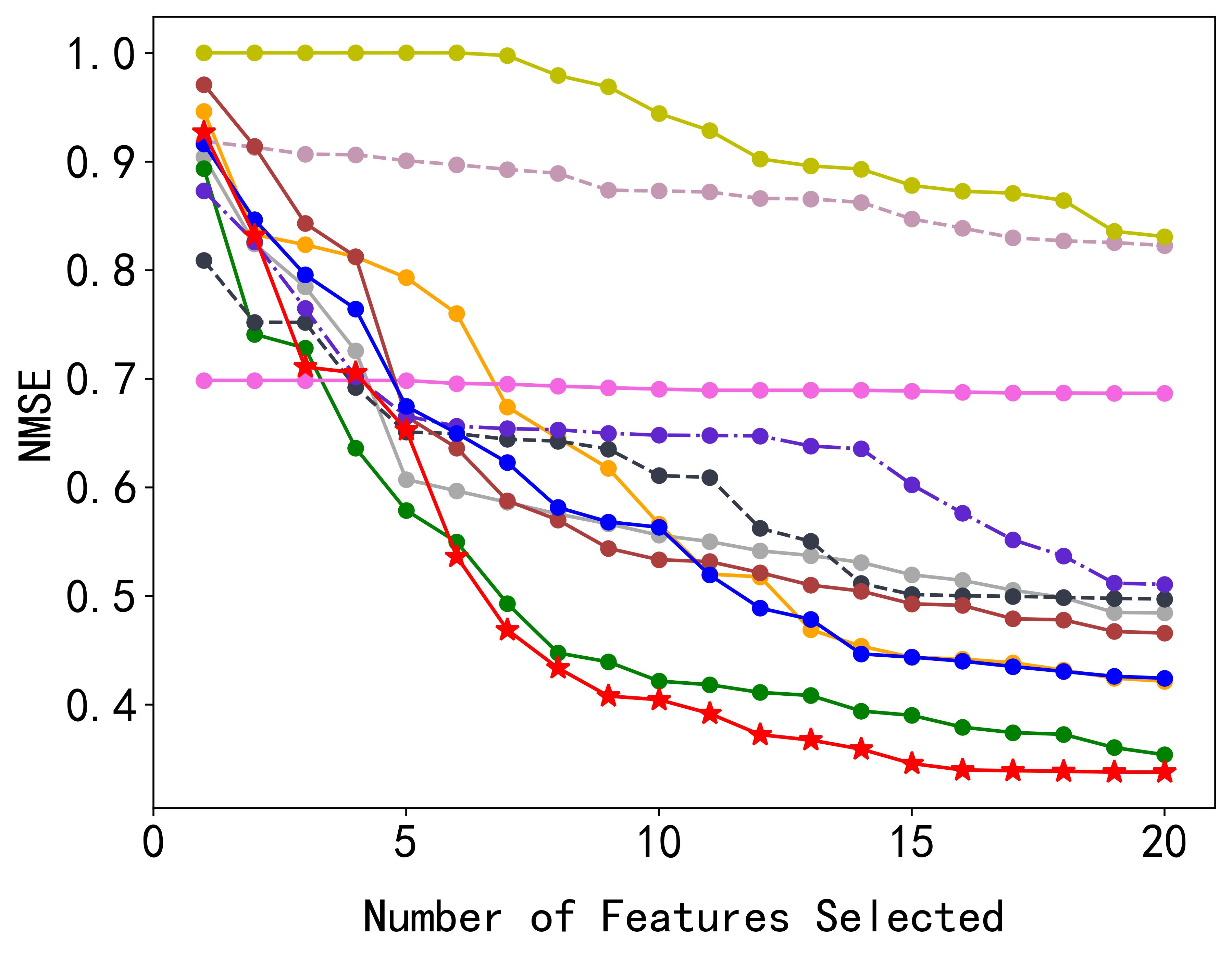}%
\label{fig_second_case}}
\hfil
\subfloat[COIL-2000]{\includegraphics[width=0.32\textwidth]{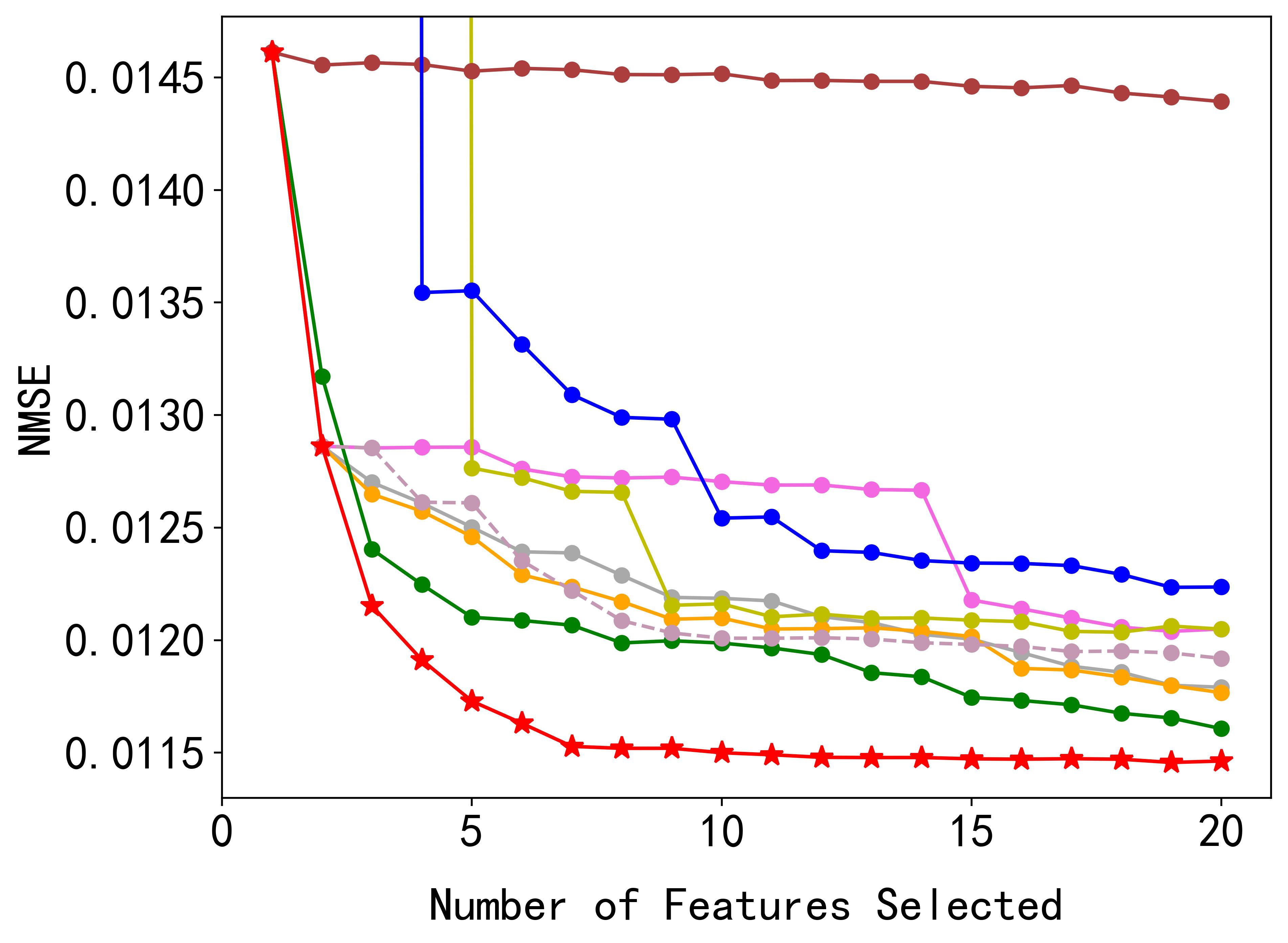}%
\label{fig_second_case}}
\hfil
\subfloat[SML]{\includegraphics[width=0.32\textwidth]{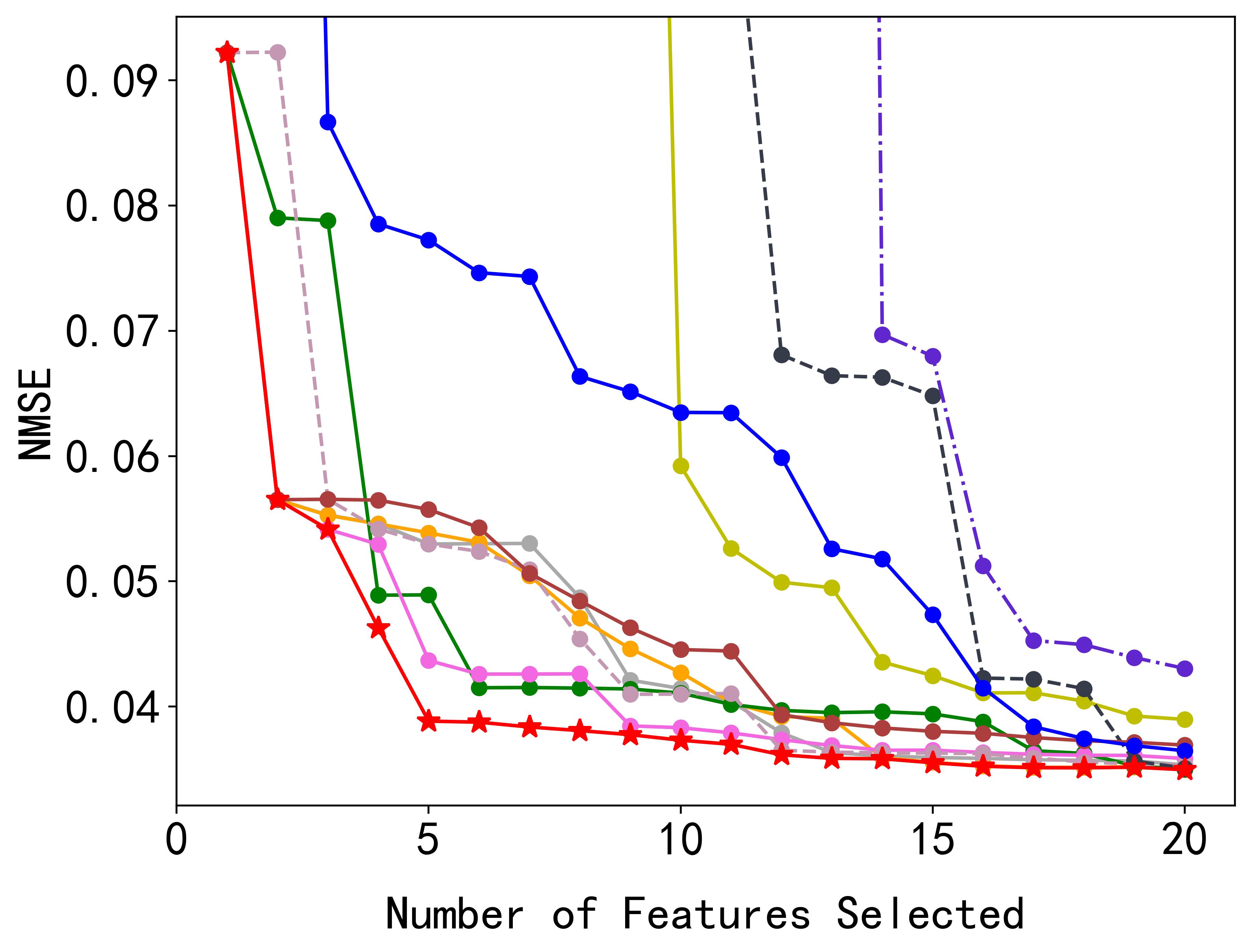}%
\label{fig_second_case}}
\hfil
\includegraphics[width=\textwidth]{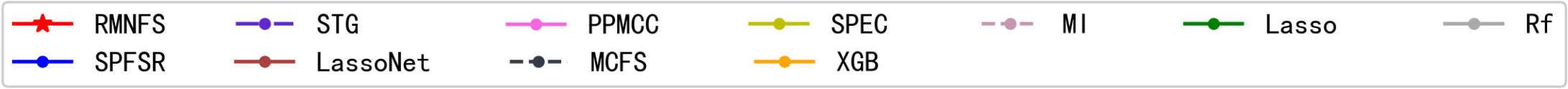}
\caption{Regression dataset results}
\label{regress_pic_res}
\end{figure*}

\section{Efficiency of Coding Redundancy}
\label{appendix_effect_redu}

\textbf{Comparison of CR and Mutual Information.}\ When measuring the redundancy of features, we replace the traditional mutual information with our proposed coding redundancy (CR) to improve computational speed and provide natural support for continuous variables. To demonstrate the effectiveness of this replacement, we randomly selected 100 features from the ORL dataset and calculated the global redundancy of each feature using both CR and mutual information. We then appropriately normalized the results, as shown in Fig. \ref{compare_to_mi}. It can be observed that the global redundancy of features obtained using our method is nearly identical to the results from mutual information calculations. Moreover, in terms of computation time, our method is only 0.192\% of the time required for mutual information, demonstrating that our proposed approach can replace mutual information to some extent when calculating global redundancy and provide a computational speedup.


\section{Regression and Classification Results}
\label{appendix_result_pic}
\begin{figure*}[htbp]
\centering
\subfloat[COIL20]{\includegraphics[width=0.3\textwidth]{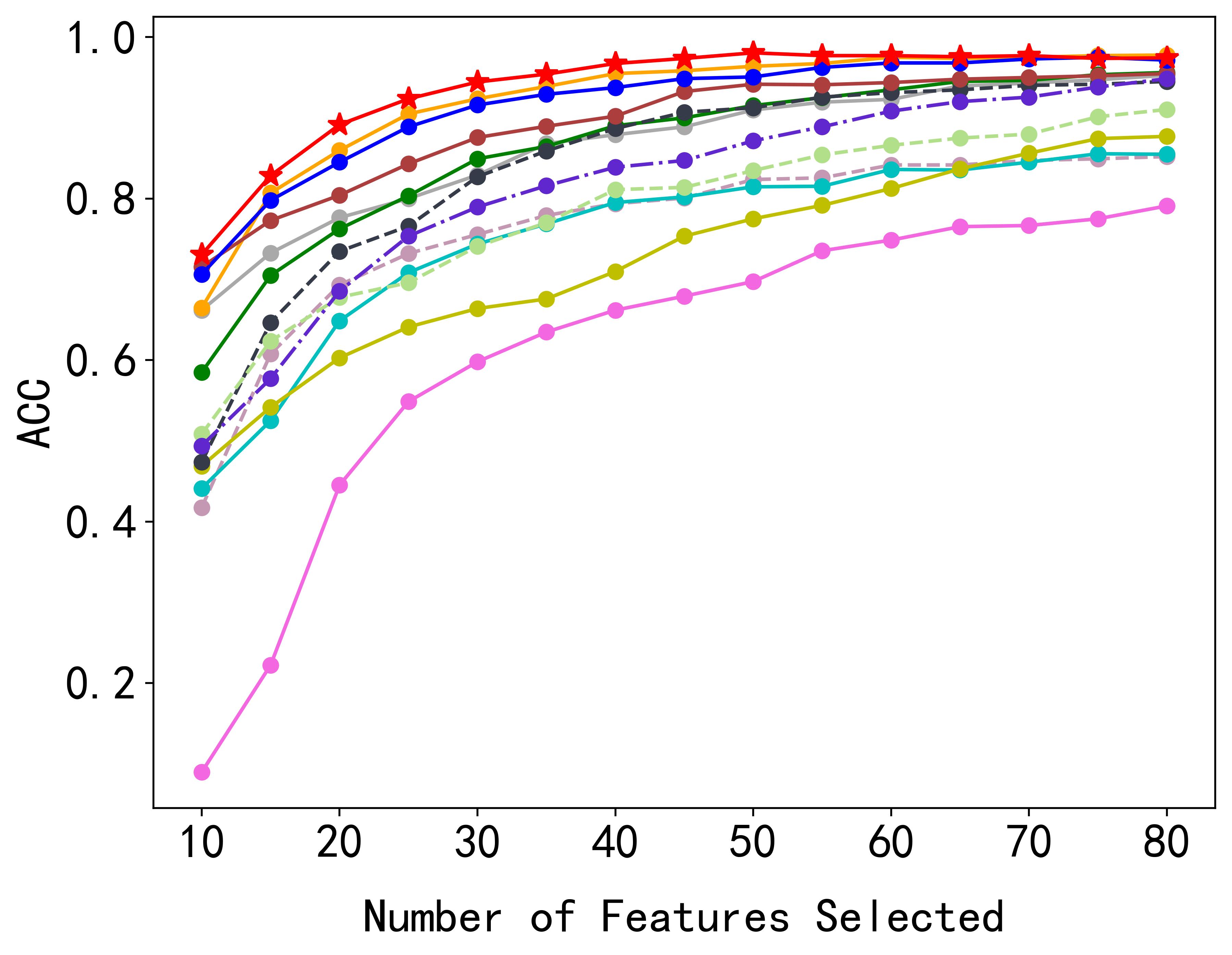}
\label{fig_first_case}}
\hfil
\subfloat[UMIST]{\includegraphics[width=0.3\textwidth]{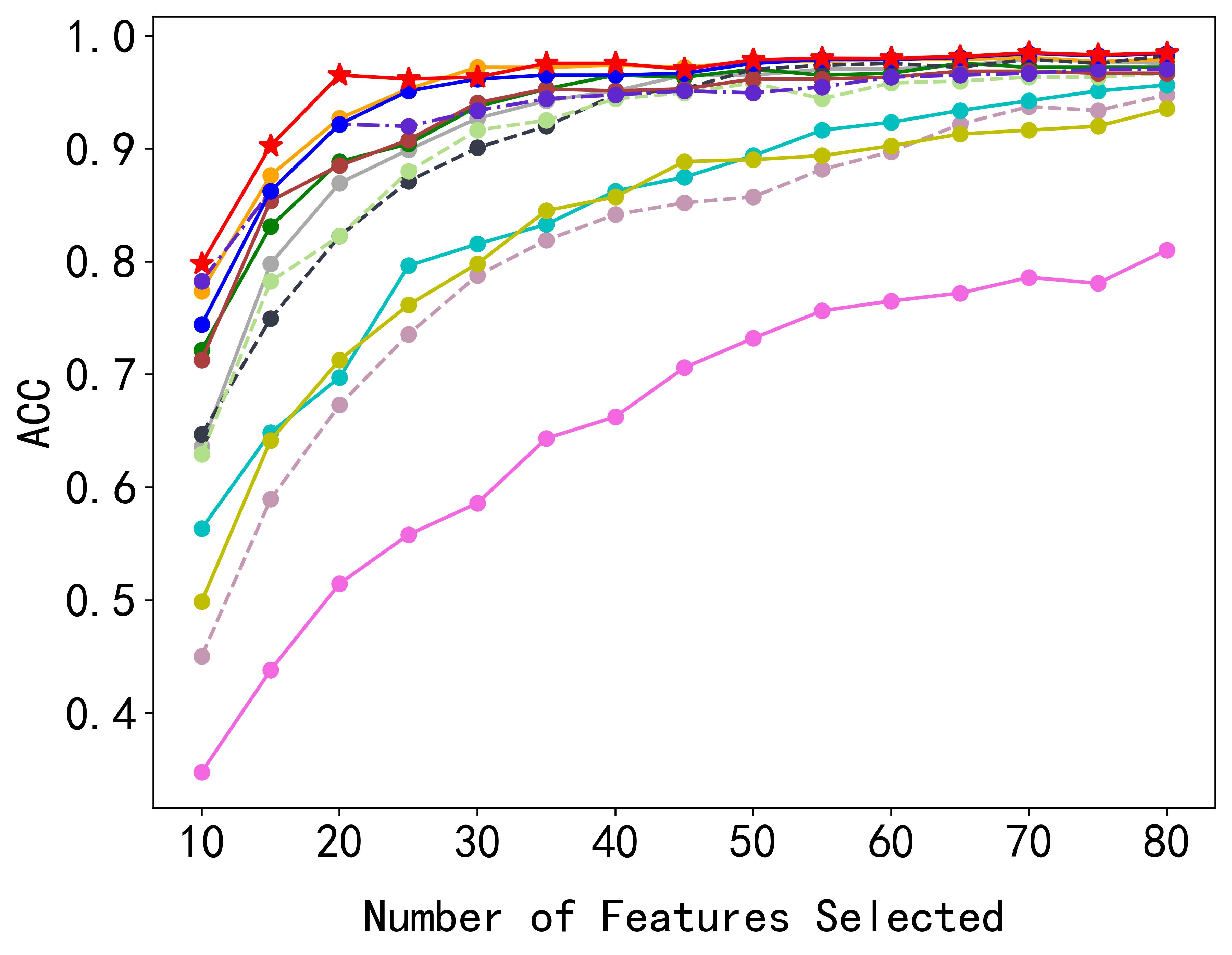}%
\label{fig_second_case}}
\hfil
\subfloat[SVHN]{\includegraphics[width=0.3\textwidth]{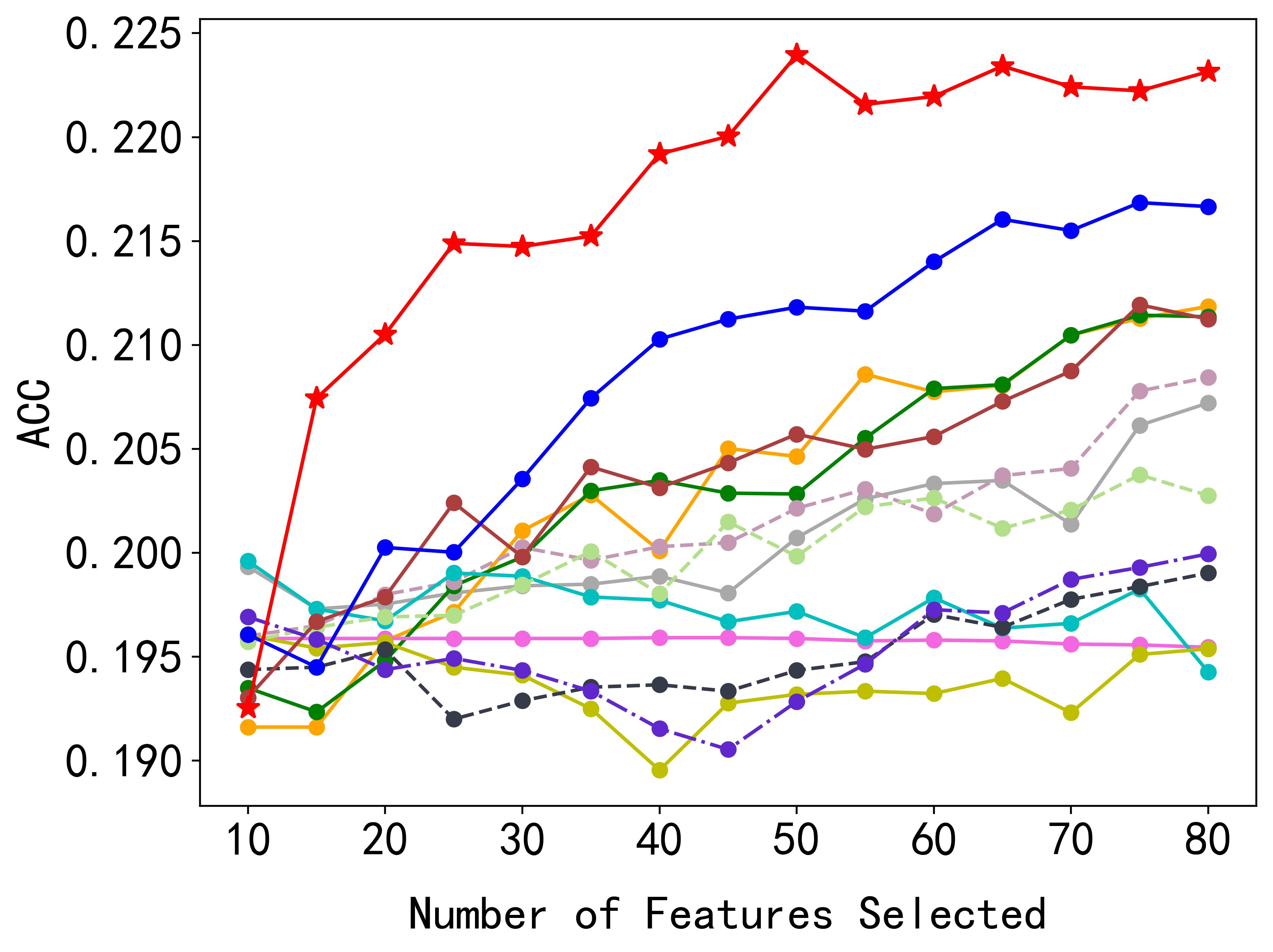}%
\label{fig_second_case}}
\hfil
\subfloat[TOX]{\includegraphics[width=0.3\textwidth]{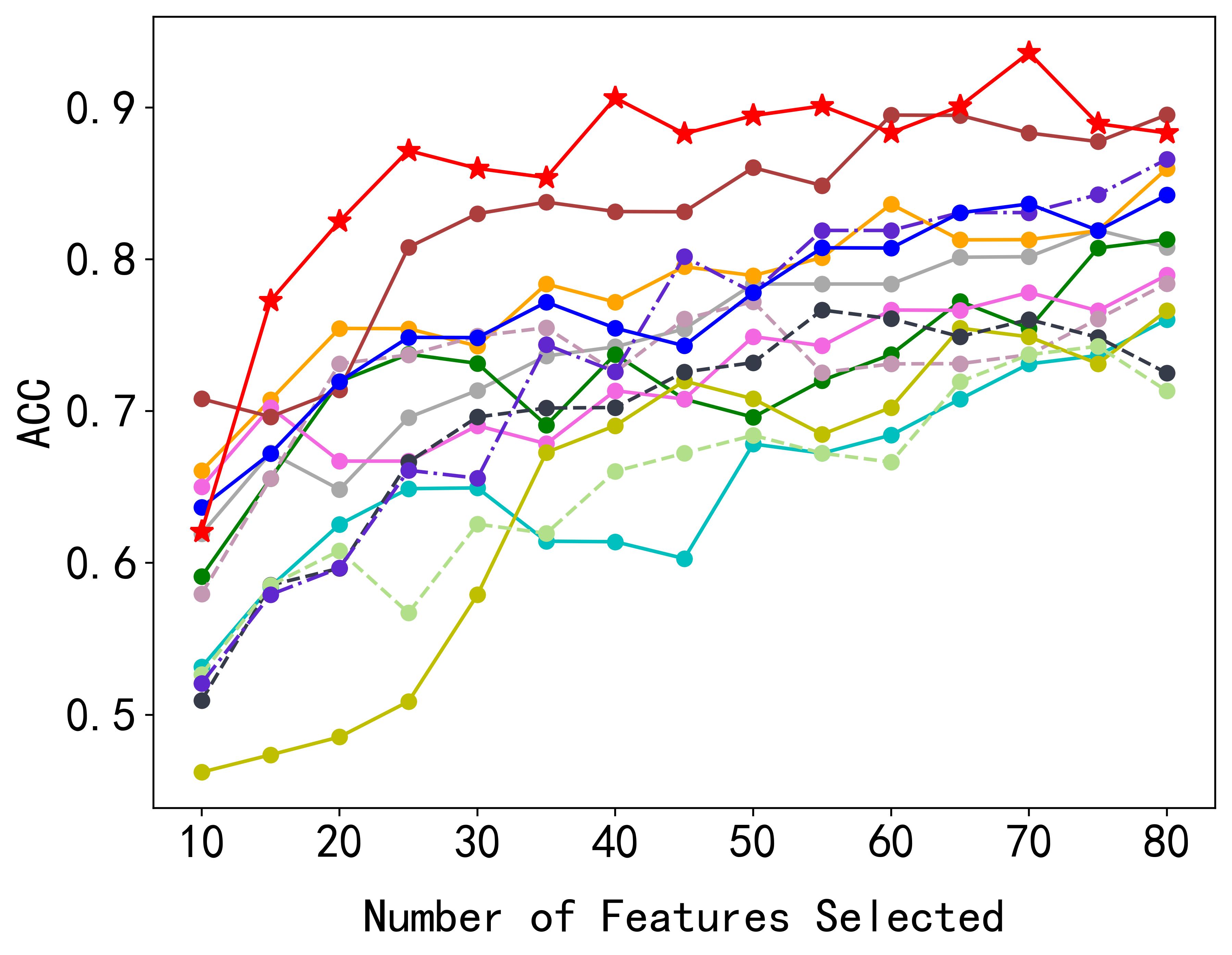}%
\label{fig_second_case}}
\hfil
\subfloat[YALE]{\includegraphics[width=0.3\textwidth]{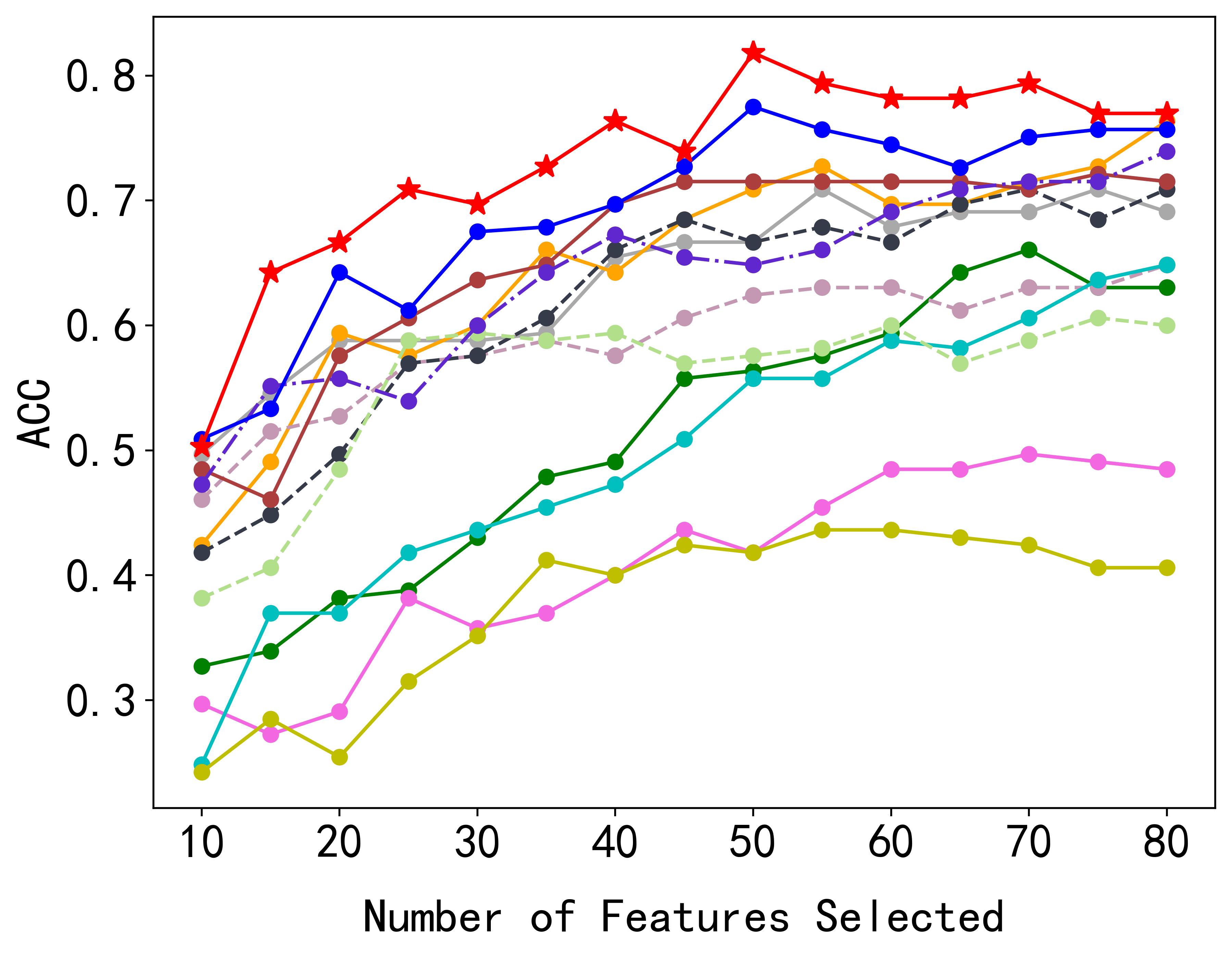}%
\label{fig_second_case}}
\hfil
\subfloat[YALEB]{\includegraphics[width=0.3\textwidth]{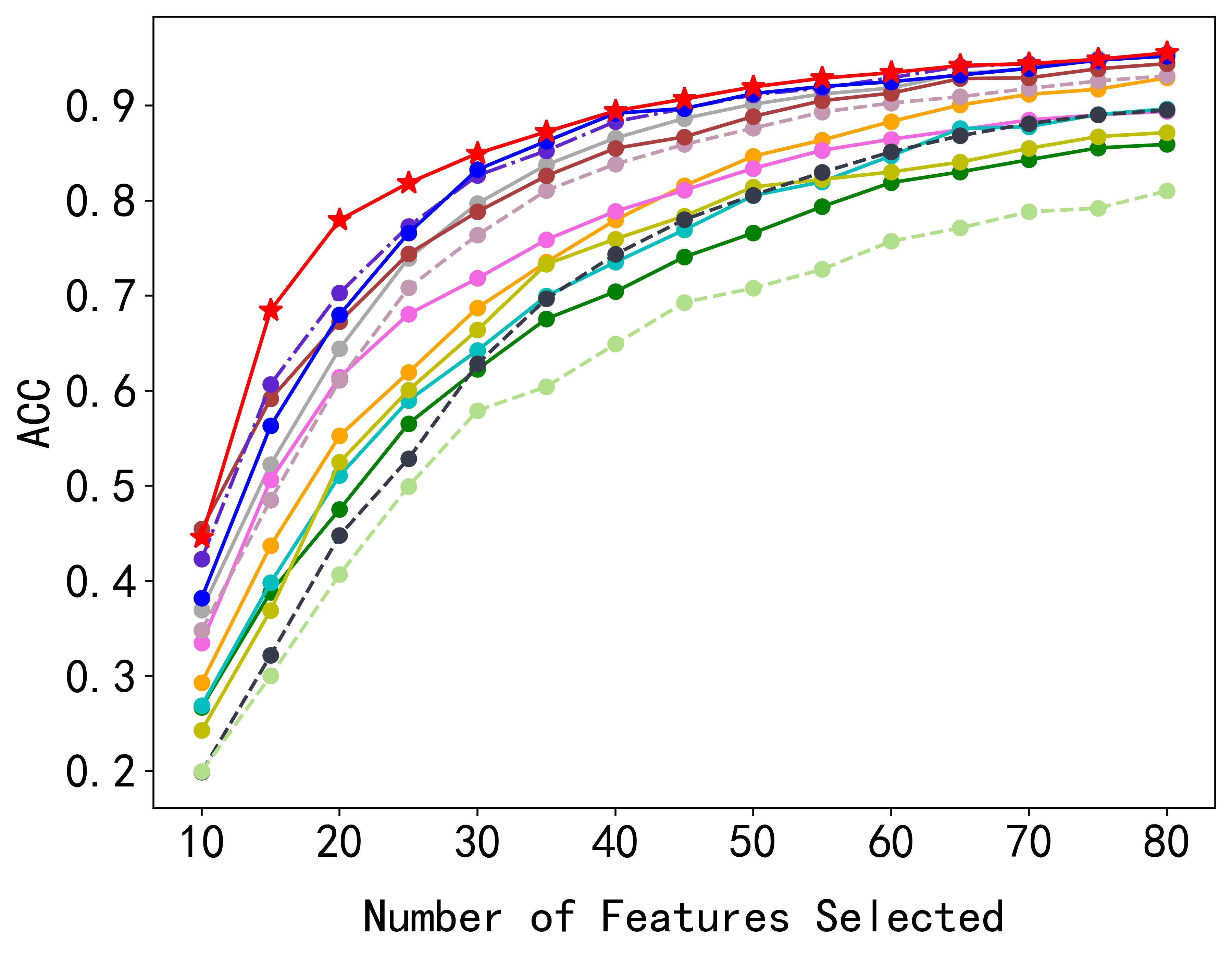}%
\label{fig_second_case}}
\hfil
\subfloat[ORL]{\includegraphics[width=0.3\textwidth]{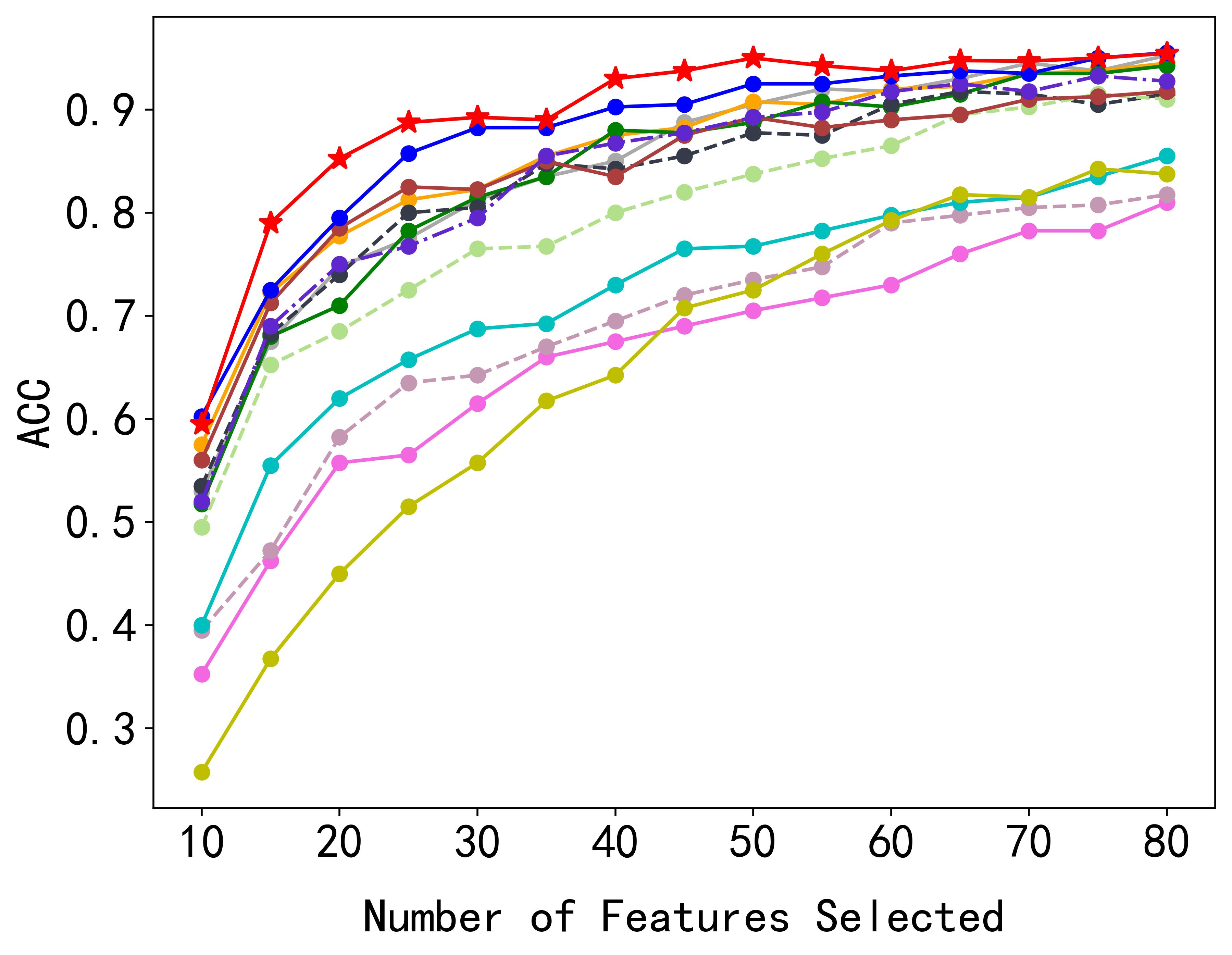}%
\label{fig_second_case}}
\hfil
\subfloat[PIE-05]{\includegraphics[width=0.3\textwidth]{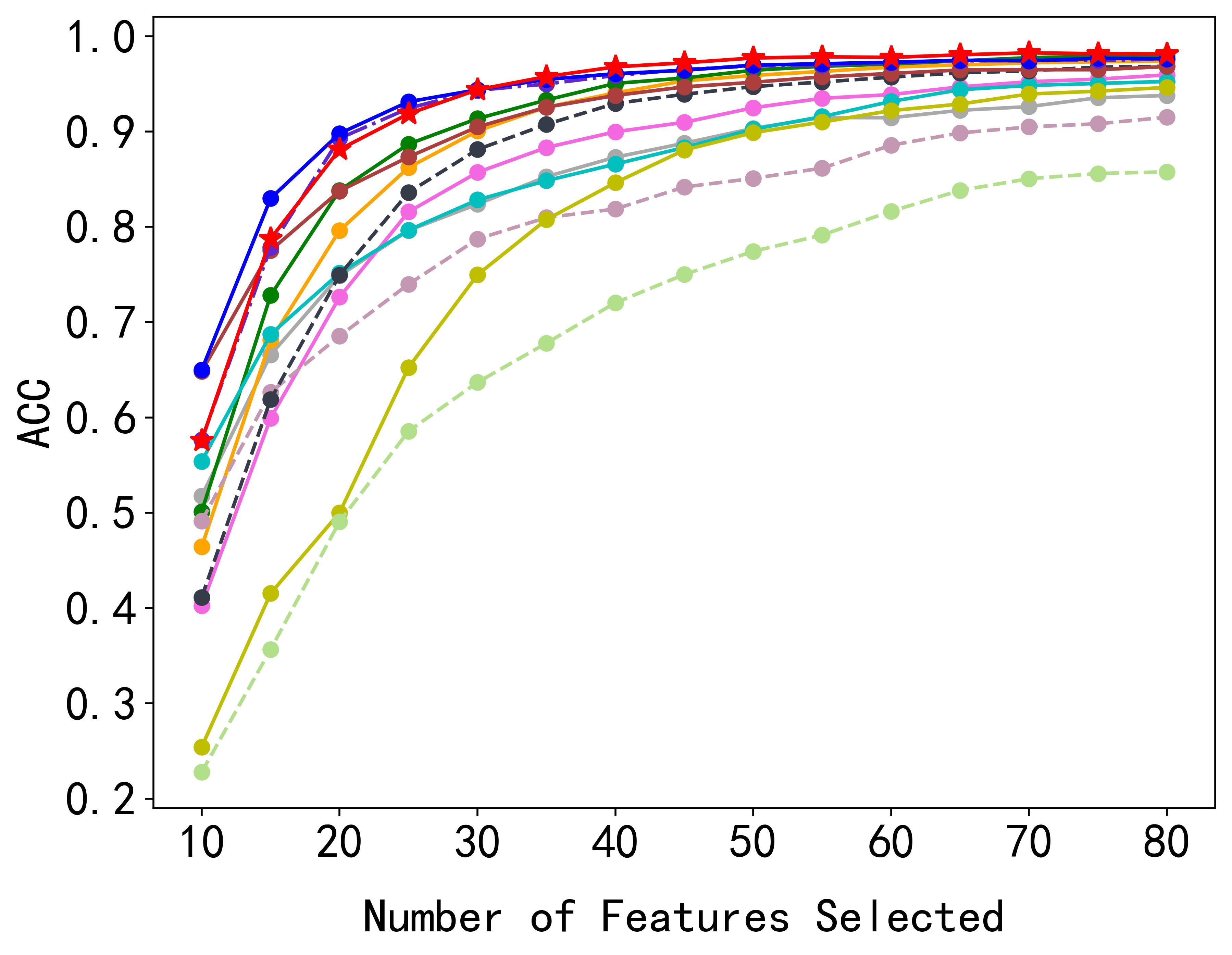}%
\label{fig_second_case}}
\hfil
\subfloat[warpAR10P]{\includegraphics[width=0.3\textwidth]{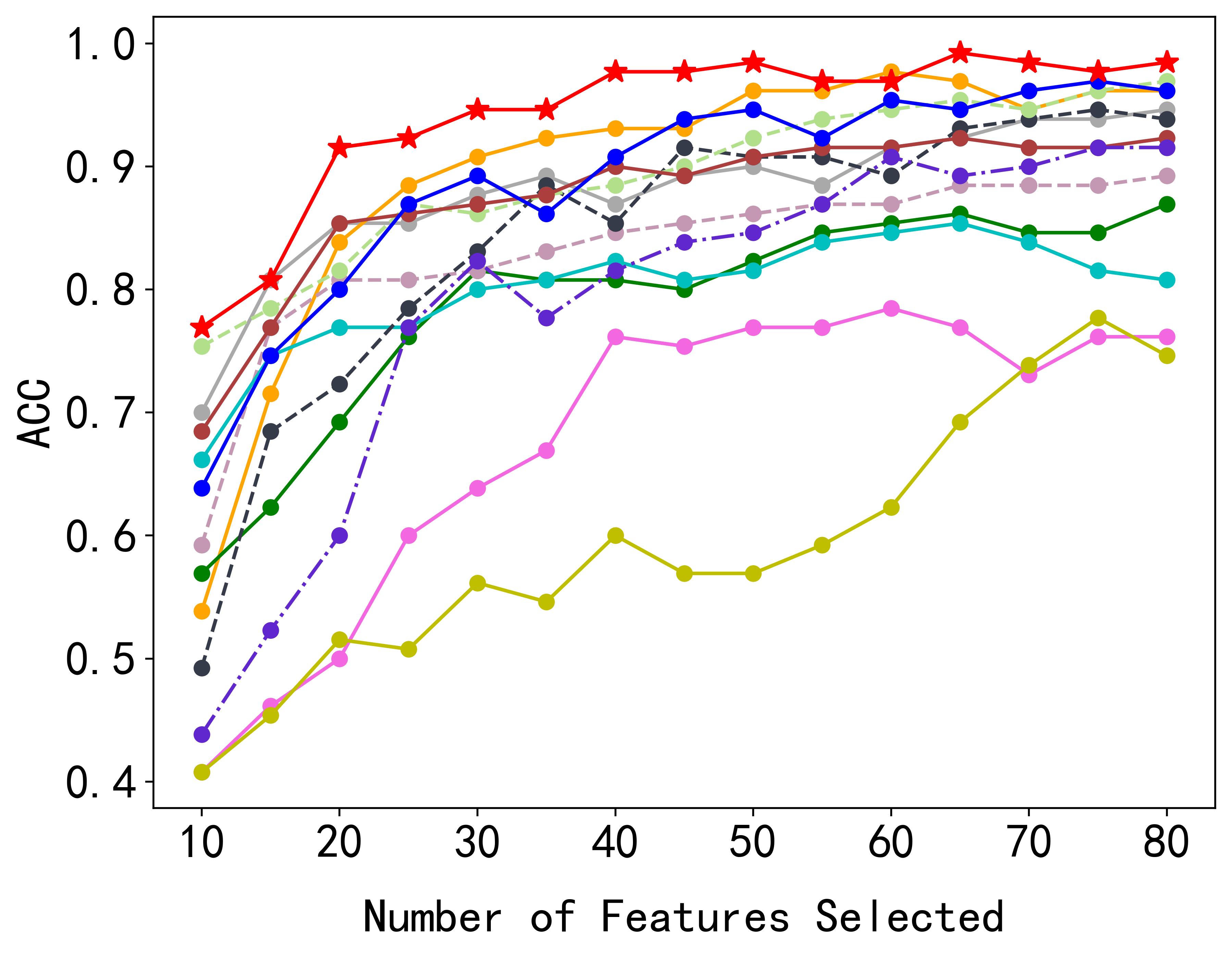}%
\label{fig_second_case}}
\hfil

\includegraphics[width=\textwidth]{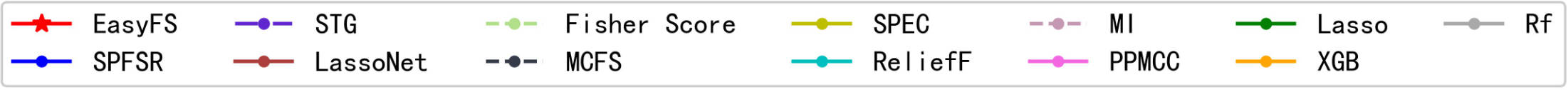}
\caption{Classification dataset results}
\label{class_pic_res}
\end{figure*}

 We depict comparison curves for all datasets in Fig. \ref{regress_pic_res} and Fig. \ref{class_pic_res}. 

\begin{figure*}[ht]
\vskip 0.2in
\begin{center}
\centerline{\includegraphics[width=0.4\columnwidth]{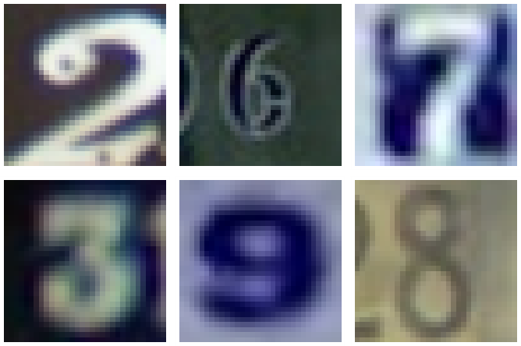}}
\caption{Example of the SVHN dataset}
\label{svhn_dataset}
\end{center}
\vskip -0.2in
\end{figure*}
\section{Visualization Analysis}
\label{appendix_vis}
\begin{figure}[ht]
\vskip 0.2in
\begin{center}
\centerline{\includegraphics[width=0.45\columnwidth]{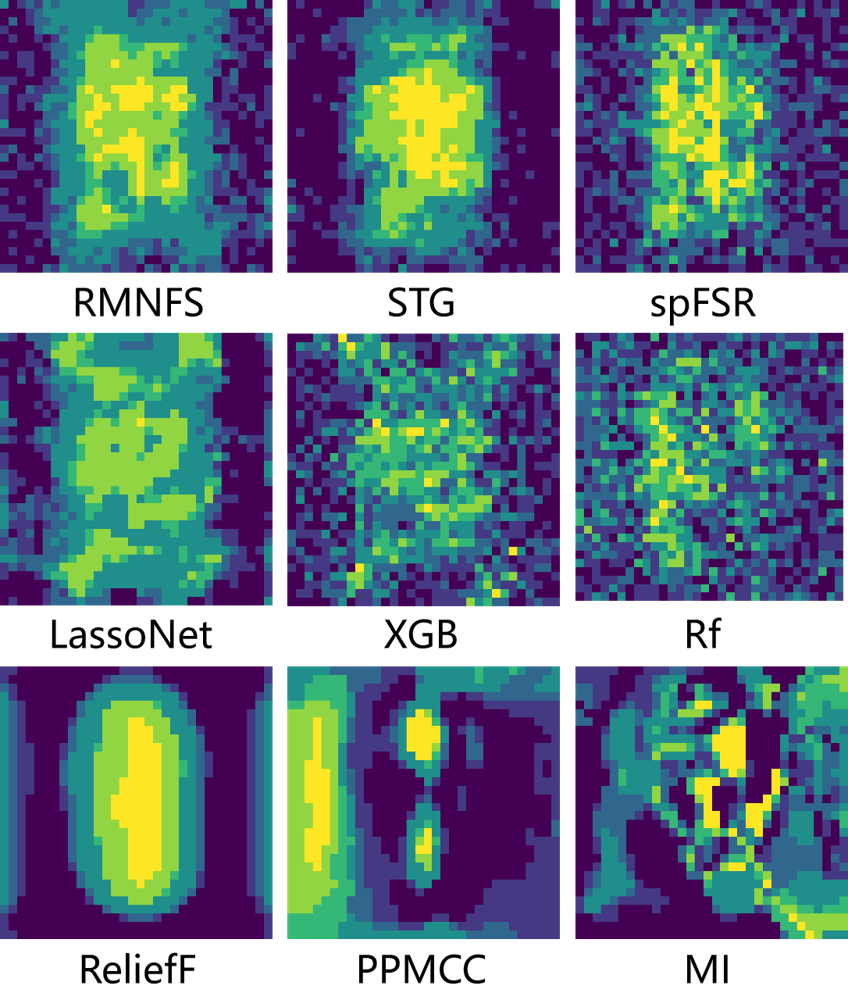}}
\caption{Visualization of results}
\label{vis}
\end{center}
\vskip -0.2in
\end{figure}

To further explore the differences in feature selection methods, we visualized the features selected by various methods in the SVHN dataset. Specifically, we ranked the features selected by each method, set the positions corresponding to the top 500 features to 255, set positions from 500 to 1500 to 127, and set the remaining positions to 0. Then, we merged the values of the three RGB channels to obtain the attention differences of the feature selection algorithms across the entire image. The results are shown in Fig. \ref{vis}. It can be observed that our method focuses the most on the middle region of the image and gradually spreads to the edges, which aligns well with human observation patterns. In contrast, the STG algorithm and Relief algorithm are overly concentrated in the middle region of the image, which can result in poor performance when dealing with digits like 2 and 7 that occupy the top and bottom edges of the image, as shown in Fig. \ref{svhn_dataset}. XGB and Bf, on the other hand, have overly dispersed focus points and do not concentrate on the high-frequency areas where relevant information is likely to be found, potentially missing important information. PPMCC appears to focus too much on regions resembling noise, leading to increased attention to image edges.

\section{Runtime Comparison}
\begin{figure}[ht]
\vskip 0.2in
\begin{center}
\centerline{\includegraphics[width=0.5\columnwidth]{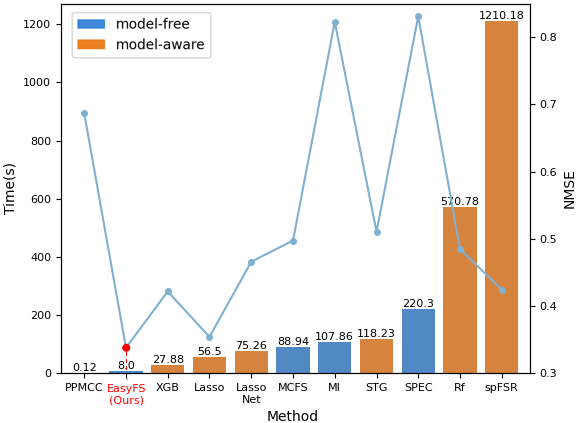}}
\caption{Runtime (by column) and NMSE (by curve) on the SLICE dataset}
\label{slice_method_time}
\end{center}
\vskip -0.2in
\end{figure}

\begin{figure}[ht]
\vskip 0.2in
\begin{center}
\centerline{\includegraphics[width=0.5\columnwidth]{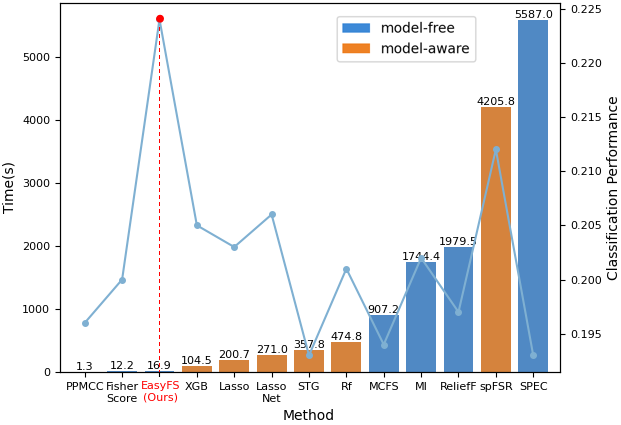}}
\caption{Runtime (by column) and Accuracy (by curve) on the SVHN dataset}
\label{svhn_method_time}
\end{center}
\vskip -0.2in
\end{figure}

To further compare the differences in runtime among different algorithms, we selected the largest datasets for experiments in both regression and classification tasks. For the regression task, the SLICE dataset was used, while for the classification task, the SVHN dataset was chosen, with approximately the top 5\% and top 2\% of features selected, respectively. The experimental results are shown in Figure \ref{slice_method_time} and Figure \ref{svhn_method_time}. It can be observed that our time advantage becomes even more pronounced on large-scale datasets.

\end{document}